\documentclass[journal]{IEEEtran}
\usepackage{enumerate}
\usepackage{multirow}
\usepackage{amsmath}
\usepackage{cite}
\usepackage{graphicx}
\usepackage{subfigure}
\usepackage{float}
\usepackage{color}
\usepackage{diagbox}
\usepackage{multirow}
\usepackage{hyperref}
\usepackage{url}
\usepackage{float}
\usepackage{lmodern}
\usepackage[T1]{fontenc}
\usepackage{pxfonts}
\usepackage{array}
\usepackage{gensymb}
\usepackage{marvosym}
\usepackage{authblk}

\begin{document}

\title{Spherical Convolution empowered FoV Prediction in 360-degree Video Multicast with Limited FoV Feedback}


\author{
\IEEEauthorblockN{Jie Li$^{1}$, Ling Han$^{1}$, Cong Zhang$^{1}$, Qiyue Li$^{2}$,  Zhi Liu$^{3}$ \textsuperscript{\Letter} \thanks{\textsuperscript{\Letter} Corresponding author} }\\
\IEEEauthorblockA{$^{1}$School of Computer and Information, Hefei University of Technology, Hefei, China\\
$^{2}$School of Electrical Engineering and Automation, Hefei University of Technology, Hefei, China\\
$^{3}$ The University of Electro-Communications,Japan\\
Email: { lijie@hfut.edu.cn,hanling@mail.hfut.edu.cn, \\zhangcong@mail.hfut.edu.cn, liqiyue@mail.ustc.edu.cn, liu@ieee.org}
}
}

\maketitle







\begin{abstract}

Field of view (FoV) prediction is critical in 360-degree video multicast, which is a key component of the emerging Virtual Reality (VR) and Augmented Reality (AR) applications. Most of the current prediction methods combining saliency detection and FoV information neither take into account that the distortion of projected 360-degree videos can invalidate the weight sharing of traditional convolutional networks, nor do they adequately consider the difficulty of obtaining complete multi-user FoV information, which degrades the prediction performance.
This paper proposes a spherical convolution-empowered FoV prediction method, which is a multi-source prediction framework combining salient features extracted from 360-degree video with limited FoV feedback information. A spherical convolution neural network (CNN) is used instead of a traditional two-dimensional CNN to eliminate the problem of weight sharing failure caused by video projection distortion. Specifically, salient spatial-temporal features are extracted through a spherical convolution-based saliency detection model, after which the limited feedback FoV information is represented as a time-series model based on a spherical convolution-empowered gated recurrent unit network. Finally, the extracted salient video features are combined to predict future user FoVs. The experimental results show that the performance of the proposed method is better than other prediction methods.
\end{abstract}

\begin{IEEEkeywords}

360-degree video, video multicast, FoV prediction, saliency detection, spherical convolution

\end{IEEEkeywords}

\section{Introduction}

Immersive multimedia, including 360-degree video and virtual/augmented reality (VR/AR) video, has recently become more and more popular as demand for interactive applications increases \cite{liu2021point,fan2019survey}. 360-degree video, also known as spherical video, is a novel type of video that can take advantage of the head-mounted displays (HMDs)  to provide an extraordinary immersive by showing the user the entire 360-degree spherical scene with his virtual position as the centre of the sphere \cite{8795062}. 
However, there is an enormous gap between the bandwidth capacity of traditional wireless technologies and the bandwidth requirements of 360-degree video streaming.

Due to the limitations of the HMD, the user can see only approximately ${12\%-15\%}$ of the entire video, which is often referred to as the field of view (FoV) \cite{zhang2020360}. Therefore, transmitting 360-degree video content in its entirety, as is the strategy of YouTube \cite{fan2019survey}, results in a large waste of bandwidth and computing resources. To provide a good experience for users, adaptive streaming techniques are proposed to optimize the delivery. Tile-based adaptive streaming is one of the most popular methods for 360-degree video transmission \cite{8957509}. It divides 360-degree video frames into multiple tiles in space and encodes each tile into multiple representations at different bitrates and quality levels. Then the transmission is optimized by adaptively controlling the video bitrate according to users' FoV and network conditions \cite{guo2018optimal, guo2021power}.

In addition, adaptive multicast over cellular networks, as proposed in \cite{8999740}, is considered a more promising way of transmitting 360-degree video. It can be implemented to support large-scale live sessions to maximize the use of wireless resources while optimizing the quality of experience for users interacting with complex 
\cite{zhai2020perceptual} , bandwidth-hungry 360-degree video. However, in some multicast  applications (e.g., VR cinema), the bandwidth of the uplink channel is much smaller than that of the downlink channel due to the asymmetry of the cellular network \cite{8281422}. This means that not all users can be supported to upload their FoV information in real time, which could otherwise lead to a feedback explosion problem \cite{rumney2013lte}. Hence, only a few users can be selected to feedback FoV information. Therefore, for 360-degree video multicast system, it is of great importance to perform FoV prediction with only a few FoV feedback streams. 

Currently, FoV prediction methods can be divided into two categories: trajectory-based and content-based. The trajectory-based approach predicts future FoVs from historical head movement trajectories of one user (single-user) or other users (cross-user). The single-user methods predict future FoV based on the historical FoVs of users \cite{7840720}, and cross-user approaches assume that different users have similar viewing behavior for the same video and predict the future FoV of the target user based on the historical FoVs of other users \cite{nasrabadi2020viewport}. However, considering only head movements of users for FoV prediction is inaccurate. Therefore, other studies \cite{9207562,8613652,9238515} have combined historical view trajectories with video object tracking to predict future FoV. For example, \cite{9207562} proposes a deep learning-based FoV prediction scheme, HOP, which jointly exploits the viewer's historical FoV trajectory and target tracking through a long short-term memory (LSTM) network. 

Content-based methods usually extract a salient feature heatmap from video content, which can be considered as the most attractive part in a video frame. Because 360-degree videos are difficult to store and process in the spherical domain, they are usually projected onto a two-dimensional (2D) plane using equal rectangular projection (ERP), cylindrical projection (CYL), or truncated pyramid projection (TPP), which leads to distortion. Such distortion of spatial variation invalidates weight sharing in convolutional neural networks (CNNs) and makes traditional deep learning-based video saliency detection methods inapplicable to 360-degree videos \cite{nguyen2018your}. One way to eliminate the distortion effect is to change the projection method. \cite{8551543} proposes to project the image in cubemap (CMP) format, where the six faces can be considered as six virtual perspective camera image planes and be processed with conventional CNNs. However, this approach does not eliminate distortion, but only minimizes its effect. Moreover, the face boundaries introduce additional discontinuities that may require subsequent processing to combine the six image planes outputs. Another approach is to change the convolution method. A CNN model with a kernel stretched to fit the shape of the patches on the ERP format is proposed in \cite{2017Flat2Sphere}, this can avoid the distortion problem to some extent. However, the shape of the filter in their scheme depends on the longitude of the patch on the sphere, leading to non-sharing of the convolution kernel parameters and making the computational and storage costs much higher.

In addition, approaches have been proposed to combine salient features with users' historical FoV information to predict future FoVs. Such as in \cite{8971933}, one user's history information is considered, which introduces uncertainty that may affect the prediction accuracy of FoV, especially in multicast scenarios. This is the case because the single user's attention may shift for a short time during the viewing procedure, resulting in historical trajectories that may not exactly match with video saliency. However, it is possible to use FoV feedback from a limited number of users combined with 360-degree video saliency detection to make more accurate FoV predictions in multiuser scenarios. 

To this end, this paper proposes a spherical convolution-based FoV prediction method, SPVP360, in 360-degree video multicast scenarios with limited FoV feedback, combining 360-degree video saliency as the main feature and several users' FoVs as supplementary features. Specifically, a SPherical Convolutional Neural Network (SPCNN) is used instead of the traditional 2D CNN for saliency detection and FoV prediction to eliminate projection error. The first step is to construct the salient feature extraction module and the convolutional block attention module (CBAM) based on SPCNN to perform saliency detection. Then a SPherical Convolutional Gated Recursive Unit (SP-ConvGRU) is used, which is a time-series model of FoV information from the FoV feedback from a small number of users. Finally, salient spatial-temporal video features and historical FoV information are combined for accurate FoV prediction. Experimental results show that the performance of the proposed method is better than other prediction methods.

In summary, the contribution of this study is threefold:

\begin{enumerate}
   \item Presenting the special FoV prediction problem in a 360-degree video multicast scenario and designing a prediction method with video saliency and a limited number of real users' FoV information.
   
   
   \item Using SPCNN to eliminate the projection distortion of 360-degree video, and proposing a saliency detection model based on SPCNN to extract spatial-temporal features from 360-degree video and introduce an attention mechanism in the network to improve the performance.
   
   
   \item Performing exhaustive experiments to show that the proposed method achieves better results than other methods on publicly available 360-degree video datasets.
   
\end{enumerate}

The remainder of this paper is organized as follows. Section \ref{sec_related_work} describes related work on 360-degree video and FoV prediction. Section \ref{sec_model} specifically introduces the proposed algorithm. Section \ref{sec_experiments} presents the experimental setup, and performance evaluation results. Finally, the conclusions of this paper are given in Section \ref{sec_conclusion}.

\section{Related Work}\label{sec_related_work}

This section presents a brief introduction to 360-degree video and its projection, and introduces work related to 360-degree video saliency detection and viewpoint prediction.

\subsection{ 360-Degree Video and Projection}
\begin{figure}[htb!]
    \centering
    \includegraphics[width=3.4in]{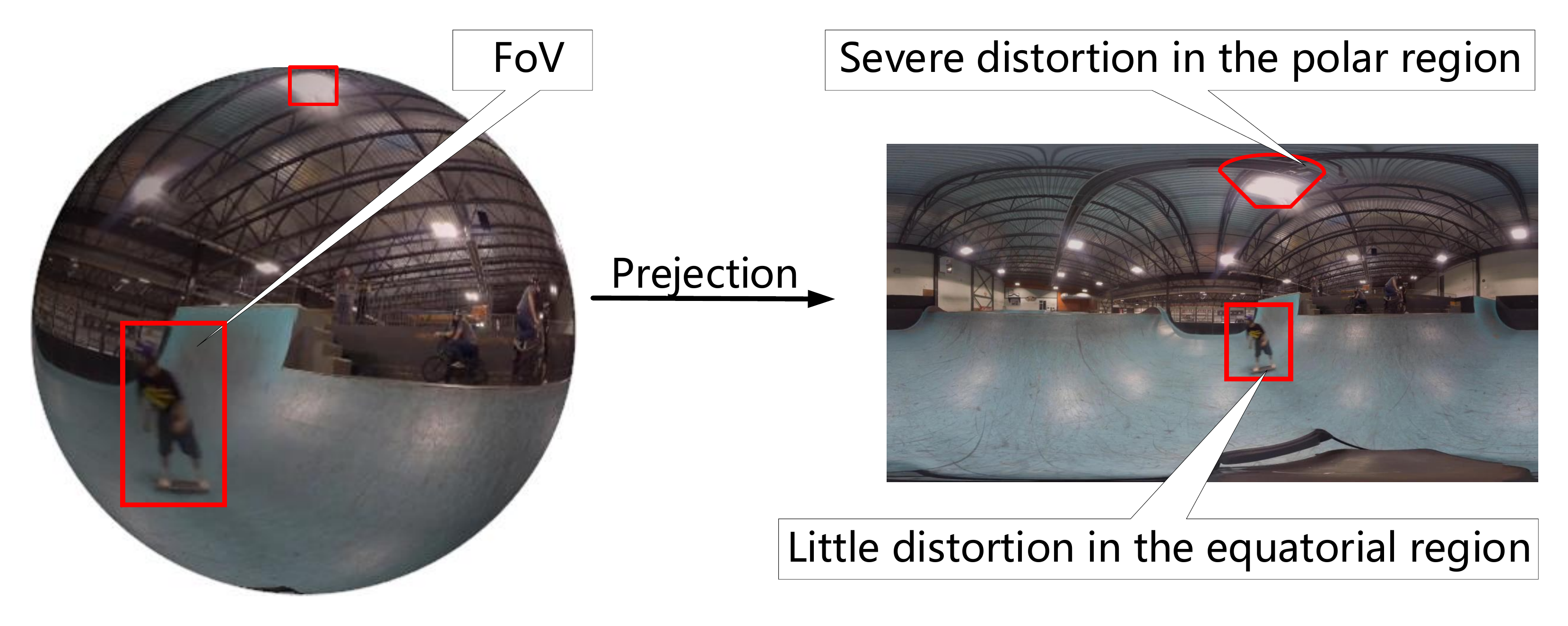}
    \caption{ Projection of a spherical image onto a 2D plane in ERP format causes distortion.}
    \label{fig_pro_erp}
     \vspace{-0.1in}
\end{figure}

A 360-degree video can be recorded by a single omnidirectional camera (e.g., a Samsung Gear 360) or aggregated and stitched using multiple cameras with separate 2D videos. To facilitate storage and processing of 360-degree video from a spherical domain, it is usually projected onto a 2D plane \cite{yu2015framework}. Current projection methods for 360-degree video include ERP, CYL, CMP, and TPP. For instance, the CYL projection maps the latitude and longitude lines as constant-spaced horizontal and vertical lines respectively to form a cylinder, which results in blanking of the north and south poles. The ERP projection directly uses the latitude and longitude on the sphere as the horizontal and vertical coordinates respectively on the original frame, which results in greater distortion in the polar region of the sphere, as shown in Fig. \ref{fig_pro_erp}. The pixels at the north and south poles are stretched to be extremely long, and the closer they are to the poles, the more severe the distortion. The CMP projection maps a spherical video to an external cube. The upper and lower faces of the cube correspond to the polar regions, and the four faces in the middle correspond to the equatorial region. At present, the most popular projection formats in panoramic video are ERP and CMP.

In addition, the HMD senses the user's viewing direction to reproject the 360-degree video in 2D format onto the sphere during 360-degree video playback. Users can enjoy spherical video content by wearing an HMD. However, due to the limitations of the HMD, the user can only see a portion of the entire video. This unique property of 360-degree video viewing provides an opportunity to reduce bandwidth consumption. Ideally, if the future FoV is known in advance, only the corresponding part of the video needs to be transmitted instead of the whole 360-degree video, which greatly reduces bandwidth consumption. Therefore, FoV prediction is of great importance as a key technology for FoV-adaptive streaming. FoV prediction is based on the fact that users tend to focus on interesting (salient) features of the scene that attract their attention. Saliency detection can reveal these features. In addition, users' FoV has a temporal meaningful association. Overall, the combination of video saliency detection and historical head motion trajectories (real-time) of users can be used to predict users' FoV in the near future. Generally, FoV prediction algorithms can be divided into two categories: trajectory-based \cite{7840720,9238515,8702654,8942328,8613652,nasrabadi2020viewport,9207562,9180528,2021PARIMA,2020Flocking,yang2020predicting} and content-based \cite{nguyen2018your,8578657,9072511,9089486,9018228,8683125,8911454,8551543,8971933,zhao2019laddernet,8779683,coors2018spherenet,2017Flat2Sphere,zhu2018prediction,8931644,ZhuYucheng2020Learning}.


\subsection{Trajectory-Based Prediction Methods}

Trajectory-based approaches \cite{petrangeli2017http,qian2018flare,yadav2020tile,7840720,8942328,nasrabadi2020viewport,9207562,8702654,8613652,9238515,9180528} predict future viewing direction from one user's (single-user) or other users' (cross-user) historical head movement trajectories. \cite{petrangeli2017http,qian2018flare,yadav2020tile,7840720} proposed to use historical head movement data to predict future FoV.
However, they only consider the user's historical viewpoint trajectory and ignore the relationship between the video target and the future FoV, resulting in poor prediction accuracy. Therefore, some studies \cite{9207562,8613652,feng2019viewport,2021PARIMA} have combined historical trajectories with video target tracking. For instance, \cite{feng2019viewport} developed a new real-time 360-degree video streaming FoV prediction method based on video content-based motion tracking and dynamic user interest modeling. 
\cite{2021PARIMA} proposed a fast and efficient online FoV prediction model, PARIMA, to predict future FoVs by using the user's past FoVs and the trajectories of the main objects as a proxy for the video content. However, there are significant differences in user trajectories in 360-degree videos. Different users may exhibit different viewing behavior patterns even for the same 360-degree video, and therefore it is insufficient to only use historical head motion trajectories to predict the FoV for all users.


\subsection{Content-Based Prediction Methods }

Content-based prediction approaches can extract salient features in 360-degree videos using saliency detection to perform FoV prediction. Saliency detection analyzes the content of video frames to find 
the areas that are most likely to attract users. Although deep CNNs have been extremely successful in traditional video saliency prediction, there has been little research on 360-degree video saliency prediction. Most existing methods \cite{Wang_2018_CVPR,Jiang_2018_ECCV,che2019gaze} directly apply techniques designed for traditional videos to 360-degree videos, but traditional deep learning-based video saliency detection methods are not applicable to 360-degree videos, because 360-degree videos projected to the 2D plane in ERP format are distorted. This distortion from spatial variation invalidates the weight sharing of traditional CNNs, resulting in poor saliency detection. For instance, \cite{Wang_2018_CVPR} proposed a video saliency model that adds an attention mechanism to the CNN-LSTM network structure to achieve fast end-to-end saliency learning. 
The CNN-LSTM is one of the latest saliency prediction methods with the best performance for 2D videos, but it is not very effective for saliency detection of 360-degree videos. Also, there are methods \cite{min2016fixation,zhu2021lavs} in which the combination of video information and audio information is proposed for saliency detection.

To eliminate the impact caused by distortion, one strategy \cite{8683125,8911454,8551543,eder2019convolutions} is to convert the 360-degree video into multiple perspective views and process them with traditional CNN on each perspective view. However, this approach does not eliminate distortion, but only minimizes its impact. For example, \cite{8911454} proposed a saliency prediction network for 360-degree videos, taking video frames and optical flows in CMP format as input, and then performing saliency prediction of these features by decoder and bidirectional convolutional LSTM. Because these views are processed separately, the face boundaries introduce an additional boundary problem that may require subsequent processing to merge the face outputs, which can affect saliency prediction performance. Another strategy is to counteract the effect of distortion by changing the convolution method, as in \cite{coors2018spherenet,9303135,2017Flat2Sphere}. For example, \cite{coors2018spherenet} proposed a new framework, SphereNet, which adjusts the sampling grid position of the convolution filter according to the geometry of the spherical image representation and wraps the filter around the sphere to avoid the effect of image distortion.

In addition, content-based prediction methods can also use the user's own past heatmap sequences with salient features for FoV prediction \cite{nguyen2018your, 8578657,9089486,9018228,9072511,
8971933,8779683,zhao2019laddernet,zhu2018prediction, 8931644, ZhuYucheng2020Learning}. 
As an illustration, \cite{8971933} proposed a multi-CNN network model that takes video frames in CMP format as input to perform corresponding saliency detection on each face separately, obtains a comprehensive saliency map using the CNN network, and then predicts the next point of view by combining the viewers' previous viewing trajectories. However, these methods only consider the historical information of a single user, which introduces uncertainty and may affect the accuracy of FoV prediction. For example, a FoV prediction method was proposed in \cite{2021Mosaic}. Unlike the existing methods, this method uses the saliency of the video and the user head tracking trajectory as the inputs for prediction, and the saliency of the video is extracted using 3DCNN, which reduces the computational requirement.
Additionally, \cite{8931644} proposed a FoV prediction model for 360-degree video that uses a spherical harmonic function to extract features in different frequency bands and different orientations to estimate saliency, and introduces visual uncertainty and balancing mechanisms in FoV prediction. 



\section{SPVP360 architecture} \label{sec_model}

This section presents an SPCNN-based method, SPVP360, which is a multi-source FoV prediction method combining salient features extracted from 360-degree videos with limited FoV feedback information. The first step is to outline the structure of the proposed network, followed by explaining the details of SPVP360 and each component. Finally, the loss function is introduced to train the network.

\subsection{Overview}

\begin{figure*}[htb!]
    \vspace{-0.05in}
    \centering
    \includegraphics[width=1\textwidth]{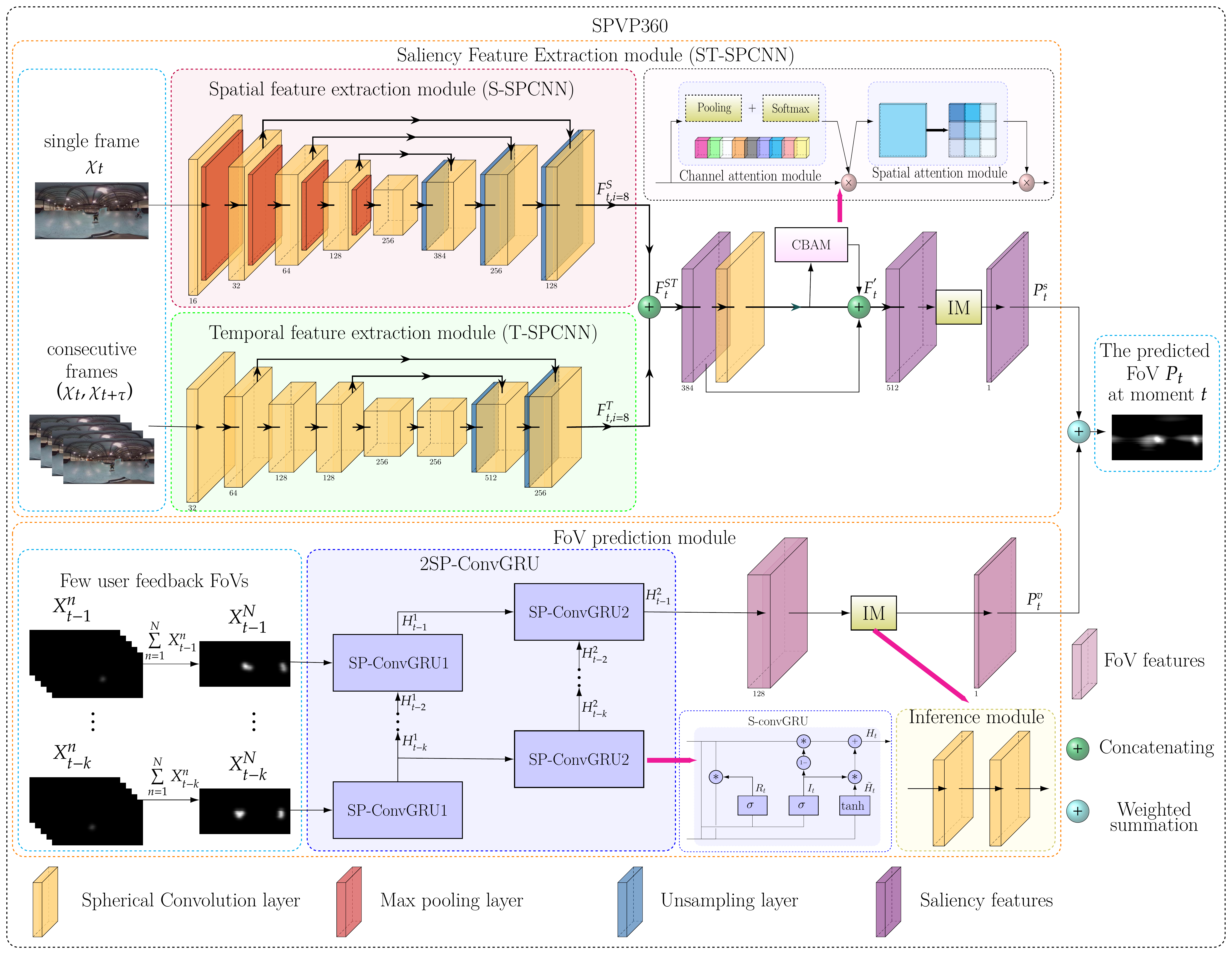}
    \caption{SPCNN-based FoV prediction network combining video saliency and limited FoV feedback.}
    \label{fig_spc_CNNflo}
     \vspace{-0.1in}
\end{figure*}

The architecture of the proposed network is shown in Fig. \ref{fig_spc_CNNflo}. Spherical convolution is used instead of traditional 2D convolution to extract salient features and FoV features in 360-degree video to improve prediction accuracy. The proposed network consists of two modules: the salient feature extraction module and the FoV prediction module. The feature extraction module takes video frames as input and extracts spatial and temporal features. The FoV prediction module takes FoV information from a small number of users as input and extracts FoV features. Accurate predictions can then be made based on video saliency and FoV features. Unlike traditional methods, the proposed network with SPCNN to eliminate the projection distortion effect uses minority FoV information combined with salient features to predict the most attractive FoV information and is suitable for 360-degree video multicast scenarios.                                                                                                                                                                                                                                                  
\subsection{Spherical Convolution on an ERP Panorama}

\begin{figure}[htb!]
    \centering
    \includegraphics[width=3.4in]{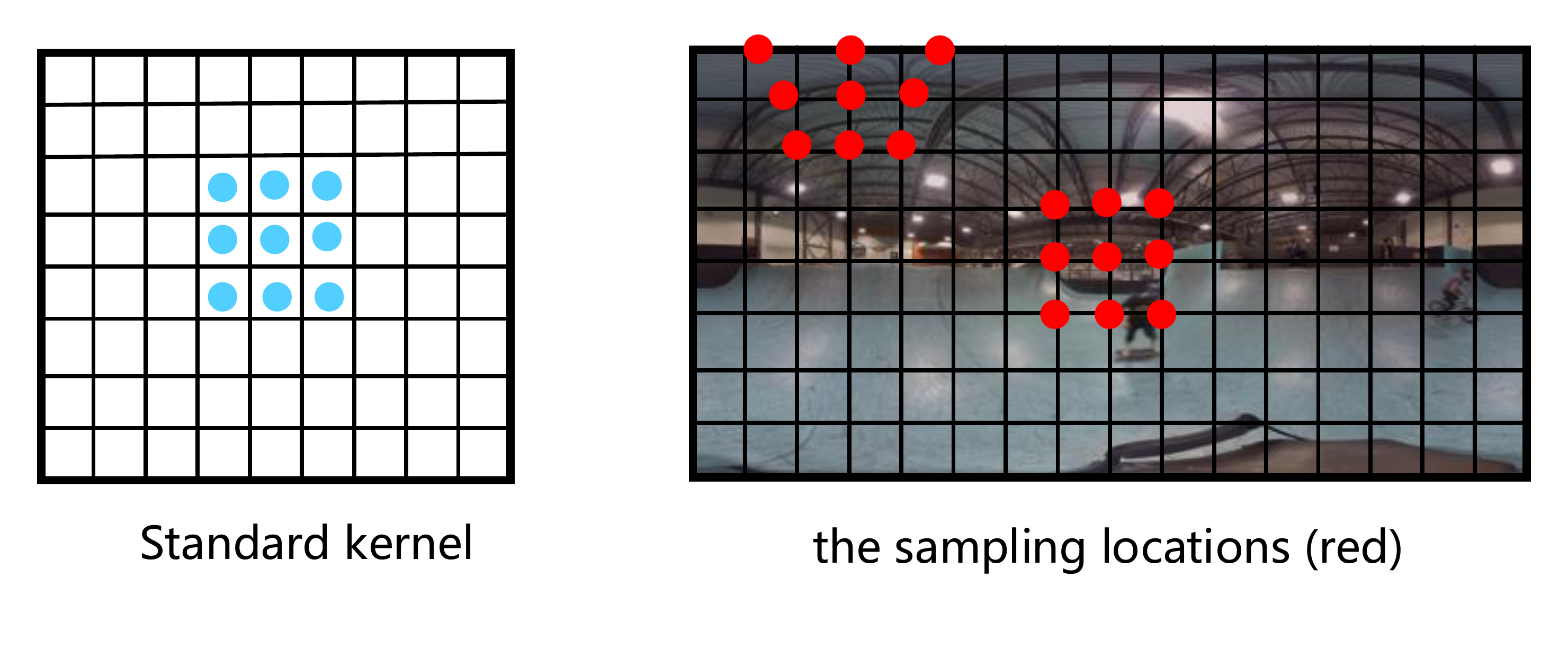}
     \vspace{-0.1in}
    \caption{Using standard convolution (e.g., with a $3 \times 3$ kernel) on a panorama in ERP format suffers from distortion of sampling locations (red) close to the poles, resulting in a failure of parameter sharing.}
    \label{fig_2d_sc}
    \vspace{-0.2in}
\end{figure}

\begin{figure}[htb!]
    \vspace{-0.05in}
    \centering
    \includegraphics[width=3.4in]{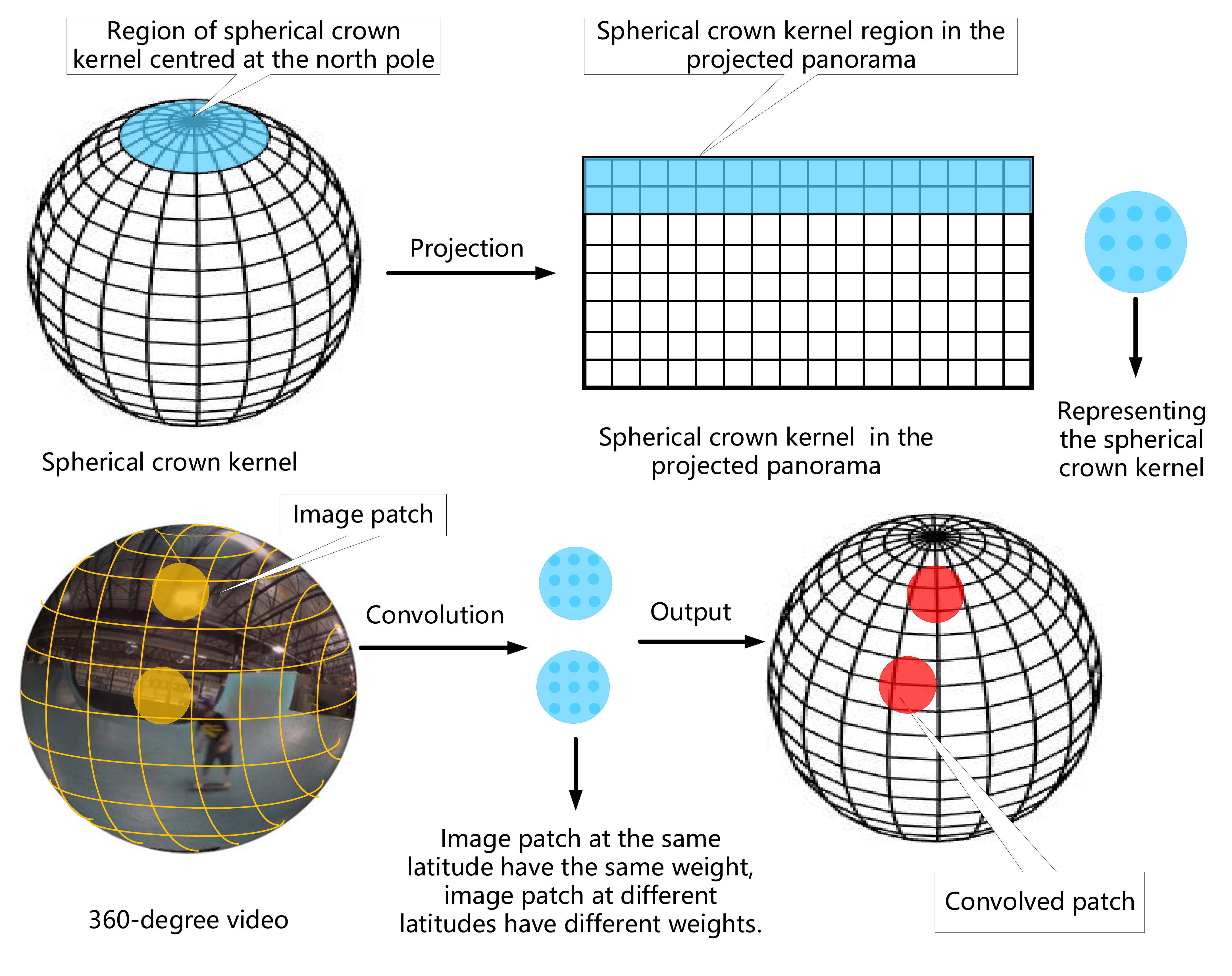} %
     \vspace{-0.05in}
    \caption{Parameter sharing. The first row shows the region of the original spherical crown kernel and the region of the spherical crown kernel in the projected panorama. The second row shows the image after spherical convolution with the same latitude parameters shared and with different latitude parameters that are not shared.}
    \label{fig_sc}
     \vspace{-0.2in}
\end{figure}

Because 360-degree video is difficult to store and process in the spherical domain, it is usually projected onto a 2D plane. Commonly used projection methods, such as ERP and CYL, cause some degree of distortion. If a 2D convolution-based neural network model is used to extract saliency in 360-degree video, the distortion of the projected image will result in low feature extraction accuracy. The reason for this is that traditional 2D CNN has only translational invariance, but no rotational invariance. Moreover, the distortion of spatial variation caused by projection invalidates the sharing of translation weights in the 2D convolutional network, as shown in Fig. \ref{fig_2d_sc}. To solve this problem, a spherical convolution \cite{driscoll1994computing} is introduced, in which a spherical kernel is defined. This kernel can be rotated and convolved with the spherical image patch. As shown in Fig. \ref{fig_sc}, the parameters of the spherical crown kernel are the same everywhere on the sphere, which implies that the kernel is shared.

Spherical convolution is defined on a spherical domain rather than the 2D image domain. It is an operation on the spherical images $f$ and the kernel filter $g$ on the sphere manifold $S^2$. To simplify the notation, the image $f$ and the filter $g$ can be modeled as a function on the sphere $S^{2} \to \mathbb{R}^{K}$, where $K$ is the number of channels and $S^{2}$ is defined as the set of points $x \in \mathbb{R}^{3}$ with norm $1$ \cite{s.2018spherical}. The shift operation on the 2D plane corresponds to the rotation operation on the spherical domain. Assuming that $R$ is a rotation matrix, a rotation can be performed using the matrix-vector product $Rx$. Obviously, the pixel value of $x$ should remain unchanged before and after rotation, that is:

\begin{equation}
    [L_Rg](x)=g(R^{-1}x)
\end{equation}

\noindent where $L_R$ is the rotation operator of the function $g$.

The proposed spherical convolution is defined analogously to the 2D version. Its essence is to multiply and sum the pixels of the spherical image and the corresponding part of the constantly rotated convolution kernel.
\begin{equation}
\label{sph_conv1}
    [f \ast g]=<L_Rg,f>={\int}_{s^{2}} \sum_{k=1}^{K}g(R^{-1}x)f(x)dx
\end{equation}
\noindent where $\ast$ denotes the spherical convolution operator.

For an ERP format projected 360-degree video, the rotation matrix $R$ can be parameterized by spherical coordinates $\lambda \in [0,2\pi]$ and $ \psi \in [0,\pi]$. Then Eq. \ref{sph_conv1} can be expressed as:
\begin{equation}
[f \ast g](\lambda, \psi)=\iint \sum _{k=0} ^K f\left(\lambda^{\prime}, \psi^{\prime}\right) g\left(\lambda^{\prime}-\lambda, \psi^{\prime}-\psi\right) \sin \lambda^{\prime} d \lambda^{\prime} d \psi^{\prime}
\end{equation}

\noindent where $f\left(\lambda^{\prime}, \psi^{\prime}\right)$ is denoted as a point on $f$ and $g\left(\lambda^{\prime}-\lambda, \psi^{\prime}-\psi\right)$ is the rotated convolution kernel. $\sin \lambda^{\prime}$ is the compensation for projection distortion. In addition, the definition has the parameter sharing property, instead of compensation for projection distortion as in traditional 2D convolution.

\subsection{Salient Feature Extraction Module (ST-SPCNN)}

The mechanism of the human visual system suggests that only small regions receive the most visual attention at high resolution, whereas other peripheral regions receive negligible attention at low resolution \cite {lin2011perceptual}. Furthermore, human attention is more likely to be drawn to the moving parts in a video frame. Considering both spatial and temporal features of 360-degree video, two sets of feature extraction models are designed: the Spatial SPherical Convolutional Neural Network (S-SPCNN) and the Temporal SPherical Convolutional Neural Network (T-SPCNN). The S-SPCNN takes one video frame as input to extract spatial features ${F}^{s}$, and T-SPCNN takes a stack of two consecutive video frames as input to extract temporal features (i.e., motion features) $F^{t}$. CBAM can then be embedded in the network to regulate its spatial-temporal features.

S-SPCNN consists of a contraction path and an expansion path, with the feature representation ${F}_{i}^{s}$ at the output of layer $i$. The contraction path resembles a traditional convolutional network, with five layers of spherical convolution and a convolution kernel of size $3 \times 3$. Each layer of spherical convolution is followed by the action of the Rectified Linear Unit (ReLU) activation layer and a maximum pooling layer of size $2 \times 2$ with a step size of $2$. The pooling layer is removed in the fifth layer to obtain a feature map with an output size of $14 \times 7$ pixels. The expansion path consists of three deconvolutional layers, each of which includes an unpooling layer, a spherical convolution layer of size $3 \times 3$, and a ReLU function. In addition, a batch normalization (BN) layer is added to each spherical convolutional layer to avoid overfitting. The final space specialization obtained is the ${F}_{t,i=8}^{S}$.

Because moving parts in a video frame are more likely to attract a user's attention, T-SPCNN is proposed to extract motion features (i.e., temporal features) of video frames. The stacked optical flow between consecutive frames $(\chi_{t},\chi_{t+\tau})$ is used as input to T-SPCNN. The structure of T-SPCNN resembles that of S-SPCNN, except that the first six convolutional layers of the T-SPCNN compressed path are spherical convolutions with a step size of $2$ and no pooling layers, and the unfolding path consists of two upper convolutional layers. Each layer undergoes BN and ReLU function actions after spherical convolution. To take full advantage of multi-scale information with different receptive domains, the input to the unfolding path is a concatenation of output features from different layers. For example, the input to the first layer of the unfolded path is the concatenation of the output features of the previous layer and the output features of the fourth layer of the contracted path.

The final output time feature is ${F}_{t,i=8}^{T}$. ${F}_{t,i=8}^{S}$ and ${F}_{t,i=8}^{T}$ are then fed into the inference module to generate spatial feature saliency map and temporal feature saliency map, respectively. Then they are concatenated to generate spatial-temporal features $F^{ST}_{t}$ of total size $384 \times 56 \times 112$, which can be expressed as:

\begin{equation}
F^{ST}_{t}=({F}_{t,i=8}^{S};{F}_{t,i=8}^{T})
\end{equation}

Attention not only tells people where to focus, but also improves the representation of interest. It can help extract important features and suppress unnecessary features. A CBAM resembling that described in \cite{Woo_2018_ECCV} is added after the spatial-temporal features $F^{ST}_{t}$ to improve saliency detection accuracy. The CBAM consists of a channel attention module (CAM) and a spatial attention module (SAM), arranged in channel-space order. The CAM uses the channel relationships between features to generate a channel attention graph. To efficiently compute channel attention $M_{c}({F^{ST}_{t}})$, the spatial information is first aggregated by averaging pooling and maximum pooling operations. Then $F^{ST}_{Avg}$ and $F^{ST}_{Max}$ are fed into the multilayer perceptron (MLP).
The output feature vectors are finally combined using the summation of elements to generate the channel attention function $M_{c}({F^{ST}_{t}})$ with a total size of $64 \times 1 \times 1$. Briefly, the channel attention is calculated as follows:
\begin{equation}
\begin{aligned} 
M_{c}({F^{ST}_{t}})&= \sigma(MLP(AvgPool(F^{ST}_{t})) + MLP(MaxPool(F^{ST}_{t})))
               \\&=\sigma(MLP(F^{ST}_{Avg}))  + MLP(M_{c}{F^{ST}_{Max}})))
\end{aligned} 
\end{equation}
where $\sigma$ denotes the sigmoid function and $F^{ST}_{Avg}$ and $F^{ST}_{Max}$ represent the average pooling and maximum pooling features, respectively. 

The SAM uses the spatial relationships between features to generate spatial attention. To compute spatial attention, two pooling operations are first performed on the spatial features to generate average pooling features and maximum pooling features. These are then concatenated, and a convolutional layer is used to generate the spatial attention $M_{s}(M_{c}^{'})$. Because the output of channel attention is the input of spatial attention, the spatial attention can be obtained as:

\begin{equation}
\begin{aligned} 
M_{c}^{'}=M_{c}({F^{ST}_{t}})\odot {F^{ST}_{t}}
\end{aligned} 
\end{equation}

\begin{equation}
\begin{aligned} 
M_{s}(M_{c}^{'})&=  \sigma(f^{7\times 7}([AvgPool(M_{c}^{'});MaxPool(M_{c}^{'})] ))
                \\&=   \sigma(f^{7\times 7}([(M_{Avg}^{'});(M_{Max}^{'})]))
\end{aligned} 
\end{equation}

\noindent where $M_{Avg}^{'}$ and $M_{Max}^{'}$ are the average and maximum pooling features after aggregating channel information, respectively. $\odot$ denotes element-wise multiplication. In this way, the channel attention values are propagated along the spatial dimension. The total size of $M_{s}(M_{c}^{'})$ is $64 \times 56 \times 112$.

The overall attention in a channel-space order process can be summarized as:
\begin{equation}
\begin{aligned} 
F_{t}^{cbam} = M_{s}(M_{c}^{'}) \odot M_{c}^{'}
\end{aligned} 
\end{equation}
 
By concatenating $F^{st}$ and the output of CBAM $F_{cbam}^{st}$, the final spatial-temporal features $F^{'}_{t}$ can be obtained:
\begin{equation}
F^{'}_{t}=({F}_{t}^{ST};F_{t}^{cbam})
\end{equation}

An inference module $A_{f}$ is constructed to generate the saliency mapping, which models the saliency of video frames based on the spatial-temporal features $F^{'}$ of the video frames. The module is a SPCNN structure consisting of a two-layer spherical convolution. Mathematically, $P^{s}$ can be computed as:
\begin{equation}
{P^{s}_{t}}=A_{f}({{F}^{ST}_{t}};F^{cbam}_{t})
\end{equation}

\begin{figure}[htb!]

    \vspace{-0.2in}
    \centering
    \includegraphics[width=3.4in]{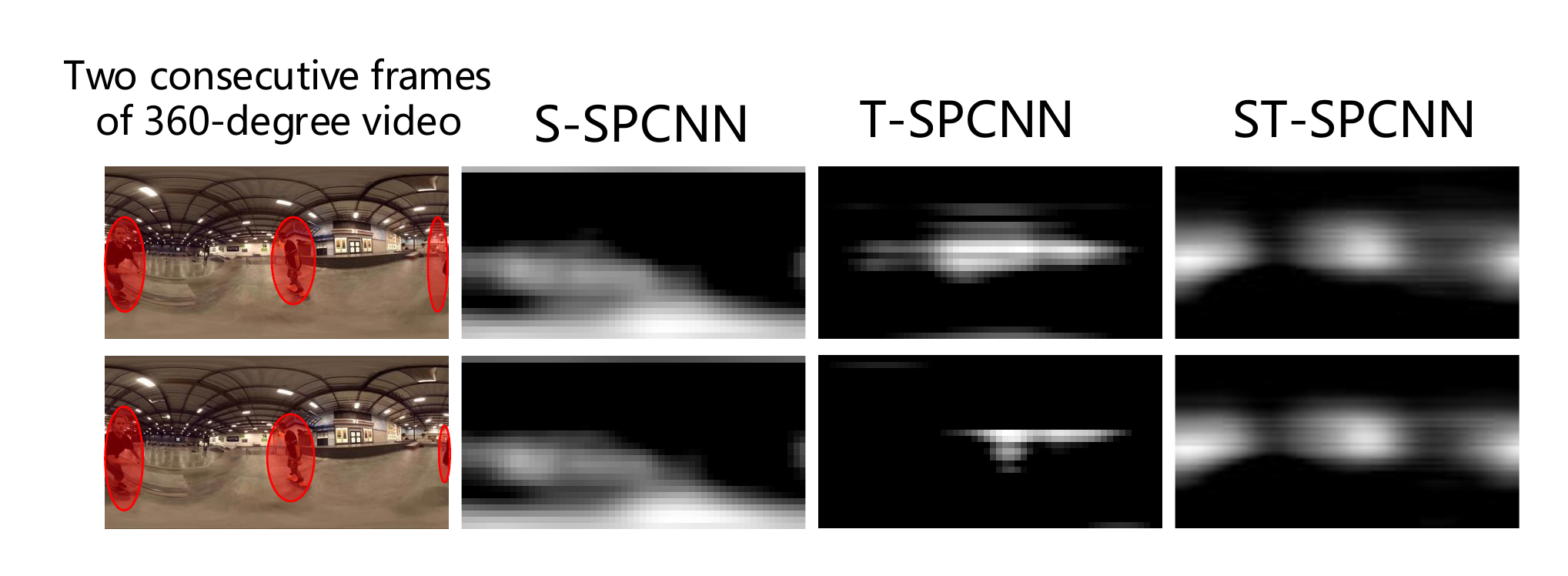}
    \vspace{-0.05in}
    \caption{Salient feature generated by the proposed extraction module. Red circles are moving parts in the video.}
    \label{fig_Inter_cnnflo}
    \vspace{-0.1in}
\end{figure}

Figure \ref{fig_Inter_cnnflo} presents results from the salient feature extraction module. Clearly, the T-SPCNN focuses only on the region with moving persons in the middle of the video frame, ignoring other salient regions, whereas the S-SPCNN has an overly large range of saliency regions and contains insignificant regions as well. In contrast, the fused spatial-temporal feature map from the proposed model is more accurate.

\subsection{FoV Prediction Module}

The model proposed in this paper combines salient spatial-temporal features of video with FoV feedback information from a small number of users to build a multi-source time-series prediction model. When users watch a video, their FoVs are not only attracted by the content of the current video frame, but also causally associated with the content of multiple frames in the previous sequence, resulting in long-distance sequencing between the FoVs of multiple frames. Therefore, the FoV information is represented from the feedback from a small number of users as a time-series model based on SP-ConvGRU, and the salient spatial-temporal features of video are combined for FoV prediction. 

During multicast transmission,, a few selected users' lines of sight can be collected from the HMD. On a spherical image, they are typically represented in the Eulerian coordinate system as $(\alpha,\beta,\gamma)$, where $\alpha$ is the cross-roll angle, $\beta$ is the pitch angle, and $\gamma$ is the yaw angle. First, these are translated into longitude and latitude:

\begin{equation}
\lambda =\sin ^{-1}\left(\frac {\beta} {\left(\alpha^{2}+\beta^{2}+\gamma^{2}\right)^{1 / 2}}\right)
\end{equation}
\begin{equation}
\psi =\tan ^{-1}(-\gamma / \alpha)
\end{equation}

\noindent where $\lambda$ and $\psi$ denote the longitude and latitude, respectively.

The user's line of sight on the ERP panorama can then be calculated as:

\begin{equation}
\mathrm{u}=(\lambda / 2 \pi+0.5) \times \mathrm{U}
\end{equation}
\begin{equation}
\mathrm{v}=(\psi /  \pi+0.5) \times \mathrm{V}
\end{equation}

\noindent where $\mathrm{U}$ and $\mathrm{V}$ are the width and height of the image, respectively. Then the FoV heatmap can be considered as a grayscale image with a 2D Gaussian distribution intensity value inside the FoV and a value of $0$ outside the FoV, which can be considered as a rectangular area centered on the line of sight\footnote{The size of the FoV depends on the specific HDM equipment. For HTC Vive, this is approximately $110 \times 90$ degrees.}.

SP-ConvGRU introduces spherical convolution, which has powerful spatial information preservation and modeling capability for temporal features, to extend traditional GRU. The element-wise multiplication $\odot$ is replaced by a spherical convolution operator in the input-to-state and state-to-state transformations of GRU. Taking the first layer of SP-ConvGRU as an example, a single SP-ConvGRU at time step $t$ can be written as follows:
\begin{equation}
I_{t}=\sigma\left(W_{z} *\left[H_{t-1}, X_{t}\right]\right)
\end{equation}
\begin{equation}
R_{t}=\sigma\left(W_{r} *\left[H_{t-1}, X_{t}\right]\right)
\end{equation}
\begin{equation}
\Tilde{H_{t}}=\tanh\left(W_{o} *\left[R_{t}*H_{t-1}, X_{t}\right]
\right)
\end{equation}
\begin{equation}
H_{t}=\left(1-I_{t} \right)*H_{t-1}+I_{t}*\Tilde{H_{t}}
\end{equation}

\noindent where $\sigma$ and $\tanh$ are the activation functions of the sigmoid and the hyperbolic tangent, respectively; $I_{t}$ and $R_{t}$ are the reset and update gates for frame $t$; and $W_{z}$, $W_{r}$, and $W_{o}$ denote the kernel weight parameters of each convolutional layer. $\Tilde{H_{t}}$ represents a memory cell, and $H_{t}$ represents the output of the hidden layer.

A two-layer SP-ConvGRU (2SP-ConvGRU) network is then constructed to predict the FoV. The numbers of feature maps for the 2SP-ConvGRU layers are both $64$. Each layer has a convolution kernel size of $3 \times 3$. 2SP-ConvGRU takes a sequence of FoVs, ${X_{t-k}^{N},\cdots, X_{t-2}^{N}, X_{t-1}^{N}}$, as input and takes advantage of the correlation of FoV information through the hidden layer. $X_{t-k}^{N}$ is obtained by $\sum_{n=1}^{N} X_{t-k}^{n}$, and $X_{t-k}^{n}$ represents the FoV of a single user $n$ at moment $t-k$. The hidden state in the second layer of 2SP-ConvGRU, $H_{t}$, is then fed into the inference module to estimate the FoV ${P}^{v}$ at time $t$.

\begin{equation}
{{P}^{v}_{t}}=A_{f}(H_{t})
\end{equation}

Finally, a fusion method based on the magnitude of global-regional disparity is used to combine the estimated FoV ${P}^{v}$ and the video saliency ${P}^{s}$. This fusion strategy approximates the human visual perception mechanism by mutual suppression of similar proximity features, which can reduce the saliency weights of the uniform distribution in salient regions and highlight local salient peaks. 
The specific steps are as follows. First, ${P}^{v}$ and $P^{s}$ are normalized to $[0,1]$. After this, global maxima ${M}_{v}$ and ${M}_{s}$ can be obtained. Next, the image is region-segmented, the extremum of each small region is calculated, and the average of all regional extrema is taken to obtain $\overline{m}_{v}$ and $\overline{m}_{s}$. Finally, the estimated FoV ${P}^{v}$ and the video saliency ${P}^{s}$ are linearly weighted together to obtain $P$.

\begin{equation}
P_{t}={{P}^{\text {s}}_{t}}*\left(M_{\text{s}}-\overline{m}_{s}\right)^{2}+{{P}^{\text {v}}_{t}} *\left(M_{\text {v}}-\overline{m}_{v}\right)^{2}
\end{equation}

By comparing the global maximum $M$ with the average ${m}_{i}$ of the corresponding local maxima, the significant features in the  significance map become more obvious and should be given more weight when the difference is large. On the contrary, when the significant features in the significance map are more uniform, they should be given less weight to suppress them when the difference is small.

Note that the proposed method can predict the user's FoV at moment $t+\varkappa$ by combining the saliency maps extracted from video frames at $t+\varkappa$ and FoV feedback at $t$, where $\varkappa$ denotes the time interval used for future prediction. In practice, $\varkappa$ can be considered as the uplink transmission delay for FoV feedback.

\subsection{Loss Definition}

Saliency detection is usually evaluated using different metrics to capture different quality factors. Commonly used metrics such as mean square error (MSE) and intersection over union (IoU) are designed for spatially uniformly discretized conventional images. Because a projected 360-degree image is non-uniformly spatially discretized, the squared error of each pixel in the panorama is weighted by its stereo angle on the sphere so that the panorama is homogeneous after discretization.

When training SPVP360, the length of the input FoV feedback sequence is set to $R$ frames. The parameters of the saliency detection model are fixed during training to extract saliency maps for the corresponding video frames and fuse them with the FoV maps.
The SPVP360 loss function can then be defined as the average MSE over $R$ frames:
\begin{equation}
D_{SPVP360}=\frac{1}{R}\sum_{ r \in R}  D_{loss}(P,G)
\end{equation}

\begin{equation}
D_{loss}(P,G)=\frac{1}{\Lambda\Psi}\sum_{\lambda \in \Lambda} \sum_{\psi \in \Psi} \omega_{\lambda,\psi}({P}_{\lambda,\psi}-G_{\lambda,\psi})^2 %
\end{equation}

\noindent where ${P}$ and $G$ indicate the output of the SPVP360 model and the ground truth, respectively. ${P}_{\lambda,\psi}$ and $G_{\lambda,\psi}$ denote the image point on the map at latitude and longitude ${\lambda,\psi}$, respectively. $\omega_{\lambda,\psi}$ is the weight of each point, which is proportional to its solid angle $\theta_{\lambda,\psi}$, defined as $\omega_{\lambda,\psi} = \theta_{\lambda,\psi} / 4\pi$ ($4\pi$ is the solid angle of the unit ball).

\section{Experiments}\label{sec_experiments}

This section describes the experimental setup and results of the proposed approach. First, the network setup is presented in the experiments, and then the proposed method is compared with other methods.

\subsection{Experimental Setup}
\begin{table}[htb!]
    \vspace{-0.1in}
    \renewcommand\arraystretch{1.25}
	\centering
	\caption{Experimental parameter settings.}
	\label{table_video_params}
	\centering
    \setlength{\tabcolsep}{5mm}{
	\begin{tabular}{c c c}
    \hline
   \multicolumn{2}{c}{SPVP360}   \\ \hline
    Initial learning rate & 0.1   \\
    Batch size & 25  \\ 
    Optimizer & SGD   \\
    Momentum & 0.9   \\
   Weight decay & $1*10^{-5}$       \\ \hline
    \end{tabular}}
    \label{table_implement_params}
\end{table}	

\textbf{Dataset and Analysis:} To train and test the proposed model, 
we use two datasets. In the first dataset \cite{Zhang_2018_ECCV}, 27 volunteers are asked to randomly watch videos with an HTC Vive HDM at a fixed starting position. This provides a benchmark for predicting the saliency of 104 panoramic 360-degree videos. We use 80 video clips for training and 24 video clips for testing. We also evaluate our model on another publicly available dataset \cite{8578657}, where the dataset contains 208 panoramic videos, each ranging from 20 seconds to 60 seconds length, with at least 31 users' viewing tracks for each video. The public datasets we use show a large diversity in content, including indoor scenes, outdoor activities, music performances, sports games, short films, etc. We classify the videos into low and high motion scenes based on the motion patterns in the videos. High sports scenes include roller skating 1, roller skating 2, basketball, etc. Low sports scenes videos include indoor scenes, outdoor scenes, etc. We number the high motion scene video (named H-video1) and the low motion scene video (named L-video1) separately.

In addition, to gain insight into the head movement of the user while watching the 360-degree video, we introduce an additional set of measures to investigate the change in FoV in successive frames. For demonstration purposes, we measure the head movement frequency by calculating the average latitude/longitude difference between two consecutive frames in the whole video and set a threshold range for longitude and latitude \cite{9238515}. If the mean longitude is greater than $ 0.65 \degree$, it is defined as a higher frequency of head movement in the horizontal direction (denoted as More). If the mean longitude is between $0.3 \degree$ and $0.65 \degree$, the head movement frequency is defined as medium frequency head movement in the horizontal direction (denoted as Middle). If the mean longitude is less than $0.3 \degree$,  the head movement frequency is defined as lower frequency head movement in the horizontal direction (denoted as Less). The definition of mean latitude in vertical direction is the same as that of mean longitude, the thresholds of which are set as $0.1 \degree, 0.3 \degree$, i.e., if the mean latitude is between $0.1 \degree$ and $0.3 \degree$, the vertical head movement frequency is defined as medium frequency of head movement (denoted as Middle). In addition, if the difference of head movement frequency in horizontal or vertical direction is less than or equal to one level, the head movement frequency is set to the higher level of the two. If the difference between the head movement in horizontal or vertical direction is more than one level, the head movement frequency is set to Middle.

\textbf{Implementation details:} 
The proposed model has been implemented in Pytorch, and the experimental code is available to the public at GitHub\footnote{ \url{https://github.com/hankebo/SPVP360}}.
The training setup is shown in Table \ref{table_implement_params}. All the experiments are performed on Nvidia Tesla V100 GPU or Nvidia Titan 2060i GPU. The sequence length used to train 2SP-ConvGRU is $3$. 
To investigate the effect of time interval, four sets of intervals $\varkappa= \left\{0.03s, 0.5s, 1s,1.5s, 2s \right\}$ are used for prediction in these experiments. This means that the estimated FoV at time $t+\varkappa$ is obtained based on FoV feedback at time $t$ and the saliency at time $t+\varkappa$.

\textbf{FoV prediction evaluation metrics:} To evaluate the performance of the proposed FoV prediction method, we use three metrics: accuracy, precision, and recall \cite{2021Mosaic}. We divide each frame into a number of small regions, each of which is called a block. Where a block pixel value of all zeros indicates a region that is not viewed by the user, and a block pixel value of not all zeros indicates a region that is viewed by the user. 
In calculating the accuracy, we use the Jaccard index. In particular, the number of blocks correctly predicted to be viewed is the number of intersection of the predicted heat map and the blocks viewed in the ground truth. 
Accuracy is the ratio of the number of blocks correctly predicted to be viewed to the number of merges between the predicted heatmap and the ground truth of the viewed blocks. Precision is a calculation of the ratio of the number of tiles to be viewed for a correct prediction to the number of tiles viewed in the predicted heat map. The higher the precision, the fewer tiles are incorrectly predicted. Recall is a calculation of the ratio of the number of tiles correctly predicted to be viewed to the number of tiles viewed in the actual heat map. The higher the recall value, the fewer tiles are incorrectly predicted not to be viewed.

\textbf{Saliency detection evaluation metrics:} To evaluate the performance of the proposed saliency detection module, ST-SPCNN, its output is compared with the ground truth and with other competitors. Three metrics are used: Normalized Scanpath Saliency (NSS), Linear Correlation Coefficient (CC), and Area Under the Curve (AUC), as proposed in \cite{8315047}. The location-based NSS metric calculates the average distance between the normalized predicted FoV and the ground truth eye gaze position, which reflects the prediction accuracy. The distribution-based CC metric represents a linear correlation between the predicted FoV and the ground truth. AUC gauges the variance between the estimated FoV and the human-annotated ground truth by calculating the true positive (TP) and false positive (FP) rates.

\textbf{Other competitors:} The proposed method is compared with five competitors, PanoSalNet \cite{nguyen2018your}, SalGAN360 \cite{8551543},  DeepVS \cite{Jiang_2018_ECCV}, GazeGAN \cite{che2019gaze} and Flare \cite{qian2018flare}. PanoSalNet includes a CNN-based panoramic saliency detection model and an LSTM-based head movement prediction model for 360-degree videos that uses transfer learning on a traditional saliency model. SalGAN360 is a saliency detection model based on generative adversarial networks, which rotates the panorama in the training set through multiple horizontal and vertical angles. DeepVS is an advanced, high-performance video saliency detection method for conventional 2D video. GazeGAN is a saliency model for 2D video-based generative adversarial networks that achieves high performance on multiple datasets.

Because SalGAN360 and DeepVS are just saliency detection models, during the FoV prediction evaluation, ablation experiments are performed to combine SalGAN360 and DeepVS separately with the proposed 2SP-ConvGRU. The resulting models are denoted as SalGAN360+ and DeepVS+. The results of the proposed ST-SPCNN saliency detection module are also illustrated and compared with those of SalGAN360, DeepVS, and the saliency detection part of PanoSalNet (called PanoSalNet-). The prediction results of 2SP-ConvGRU (named Viewport) are also compared with SPVP360. In addition, we add Flare as an experimental comparison, which is a simple method to predict future FoV using historical head motion data through machine learning.

\subsection{Experimental Results}

\begin{figure}[htb]
\centering
\subfigure[]{
   \begin{minipage}[t]{1\linewidth}
			\centering
			\vspace{0pt}
			\includegraphics[width=1\textwidth]{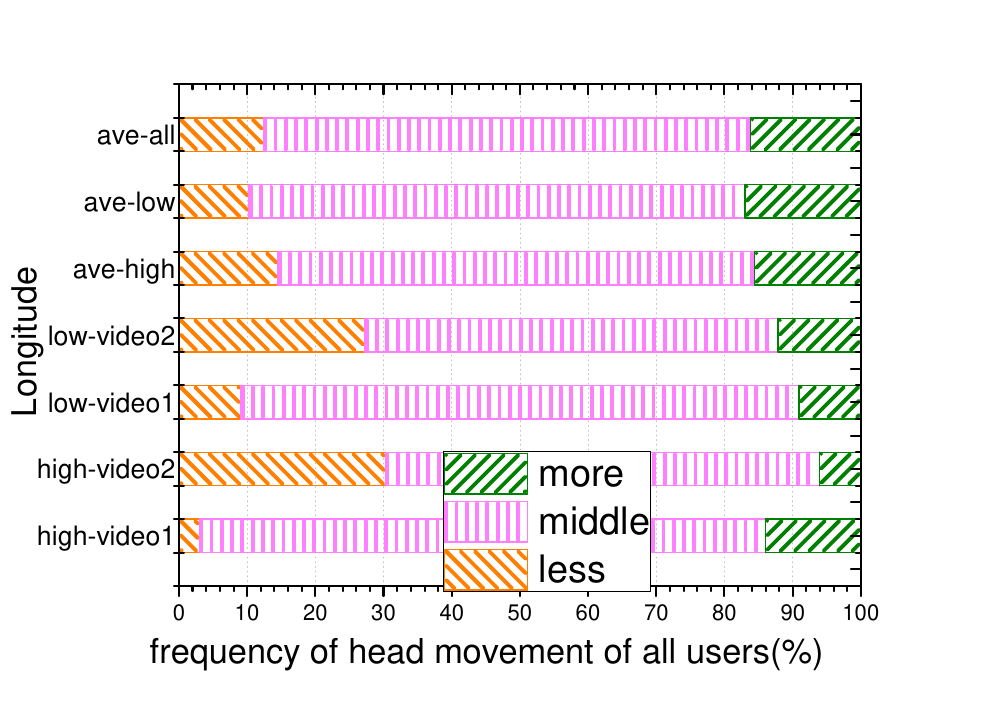}
    \end{minipage}%
\label{fig_head_lo}
}

\vspace{-0.3in}
\subfigure[]{
 \begin{minipage}[t]{1\linewidth}
			\centering
			\vspace{0pt}
			\includegraphics[width=1\textwidth]{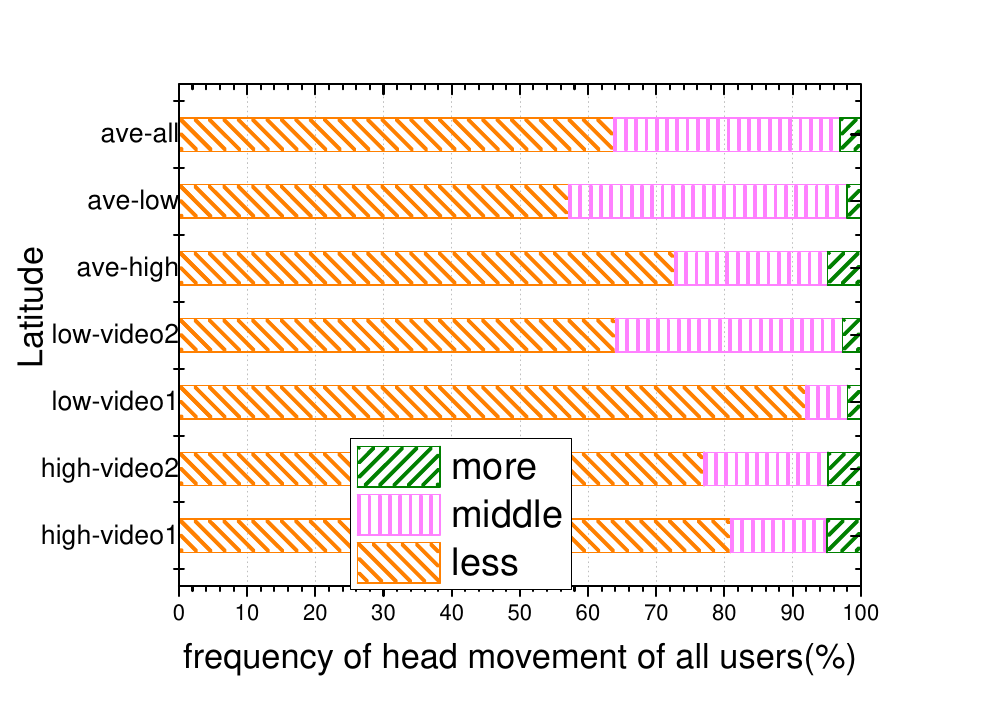}
 \end{minipage}
\label{fig_head_la}
}
\vspace{-0.1in}
\caption{ The distribution of head movement frequency.}

\label{fig_head_ana}
\vspace{-0.1in}
\end{figure}

\textbf{The Results of head movement frequency analysis:} The frequency of user's head movements in different types of videos is analyzed in the experiments. In Fig. \ref{fig_head_ana} the high-video and low-video represent the distribution of head movement frequencies of all users for that particular single video, respectively. The ave-high and ave-low represent the average distribution of user head movement frequencies for all high motion type videos and low motion type videos, respectively. The ave-all represents the average distribution of head movement frequencies of users for all videos. From the Fig \ref{fig_head_lo}, we can see that most users' head movement frequency is at Middle. In addition, users tend to have more significant head movement in horizontal direction and less frequent head movement in vertical direction, partly because prolonged head down and head up can make users feel uncomfortable.

\begin{figure}[htb]
\centering
\vspace{-0.2in}
\subfigure[]{
   \begin{minipage}[t]{1\linewidth}
			\centering
			\vspace{0pt}
			\includegraphics[width=1\textwidth]{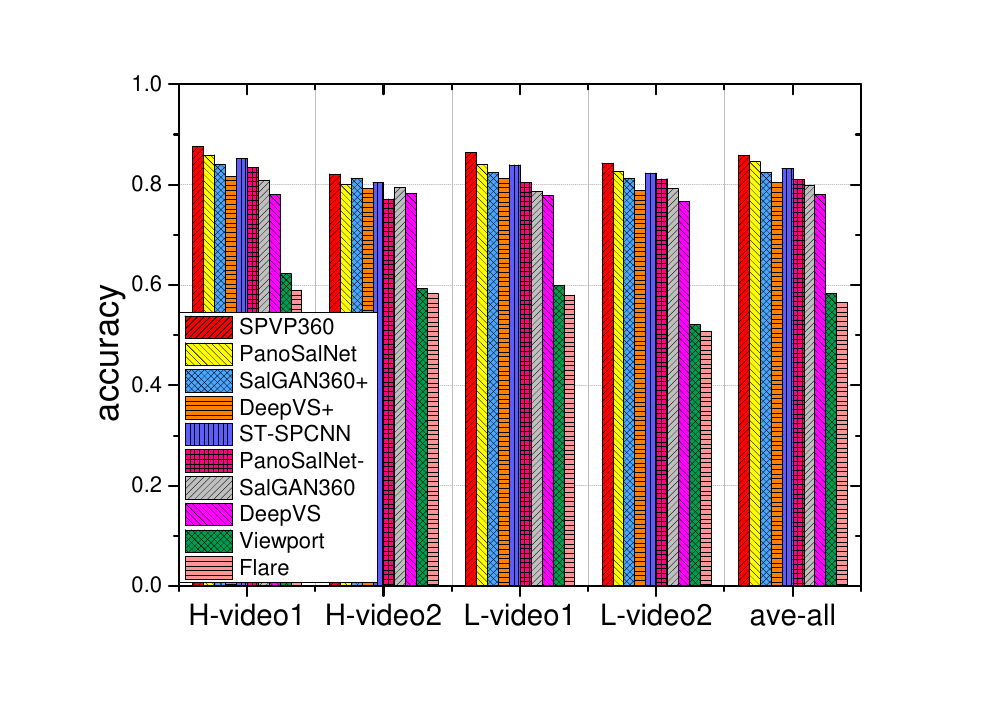}
    \end{minipage}
\label{fig_vp_vall_acc}
}
\vspace{-0.3in}

\subfigure[]{
 \begin{minipage}[t]{1\linewidth}
			\centering
			\vspace{0pt}
			\includegraphics[width=1\textwidth]{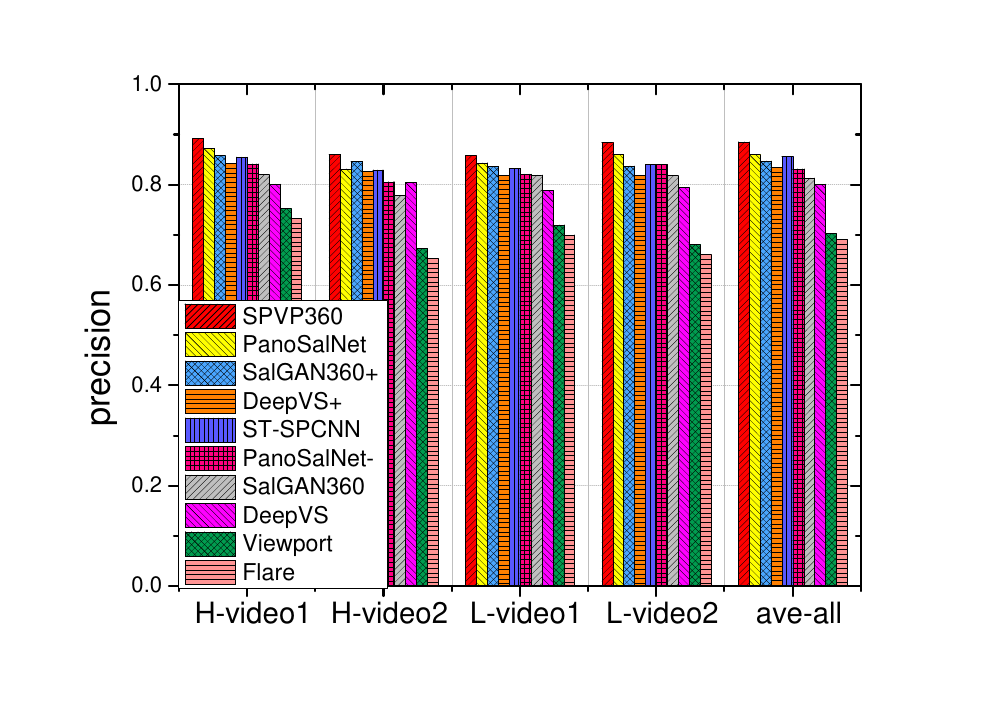}
 \end{minipage}
\label{fig_vp_vall_pre}
}

\vspace{-0.3in}
\subfigure[]{
 \begin{minipage}[t]{1\linewidth}
			\centering
			\vspace{0pt}
			\includegraphics[width=1\textwidth]{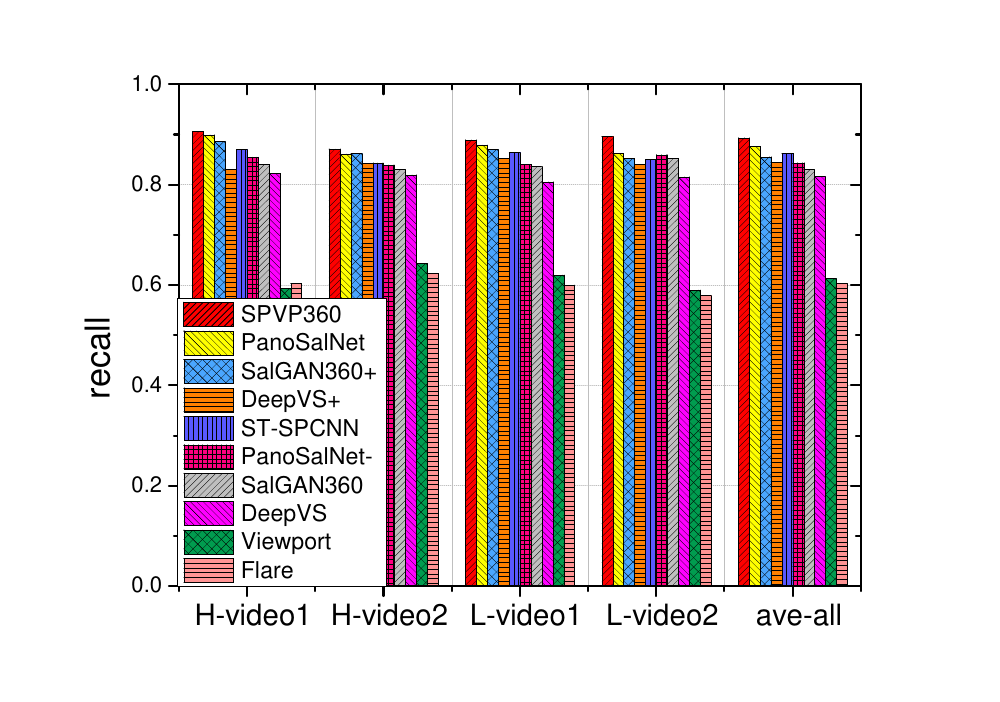}
 \end{minipage}
\label{fig_vp_vall_re}
}
\vspace{-0.1in}
\caption{The performance comparison for FoV prediction in dataset \cite{Zhang_2018_ECCV}.}
\vspace{-0.1in}
\label{fig_vp_vall}
\end{figure}

\begin{figure}[htb]
\centering
\vspace{-0.3in}
\subfigure[]{
   \begin{minipage}[t]{1\linewidth}
			\centering
			\vspace{0pt}
			\includegraphics[width=1\textwidth]{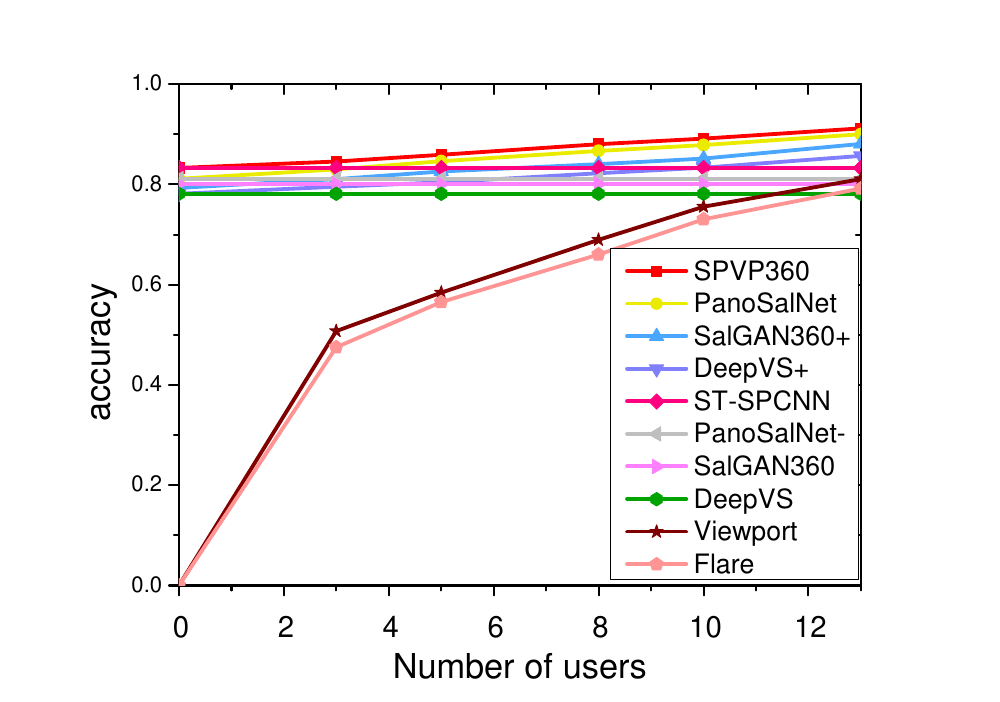}
    \end{minipage}
\label{fig_vp_num_acc}
}

\vspace{-0.3in}
\subfigure[]{
 \begin{minipage}[t]{1\linewidth}
			\centering
			\vspace{0pt}
			\includegraphics[width=1\textwidth]{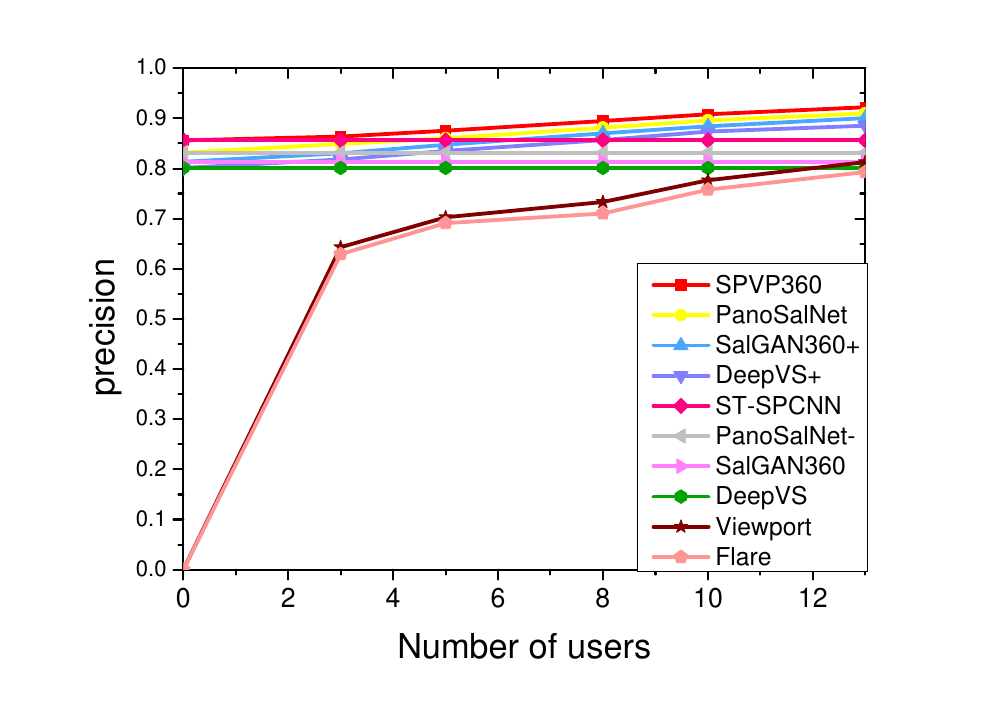}
 \end{minipage}
\label{fig_vp_num_pre}
}

\vspace{-0.3in}
\subfigure[]{
 \begin{minipage}[t]{1\linewidth}
			\centering
			\vspace{0pt}
			\includegraphics[width=1\textwidth]{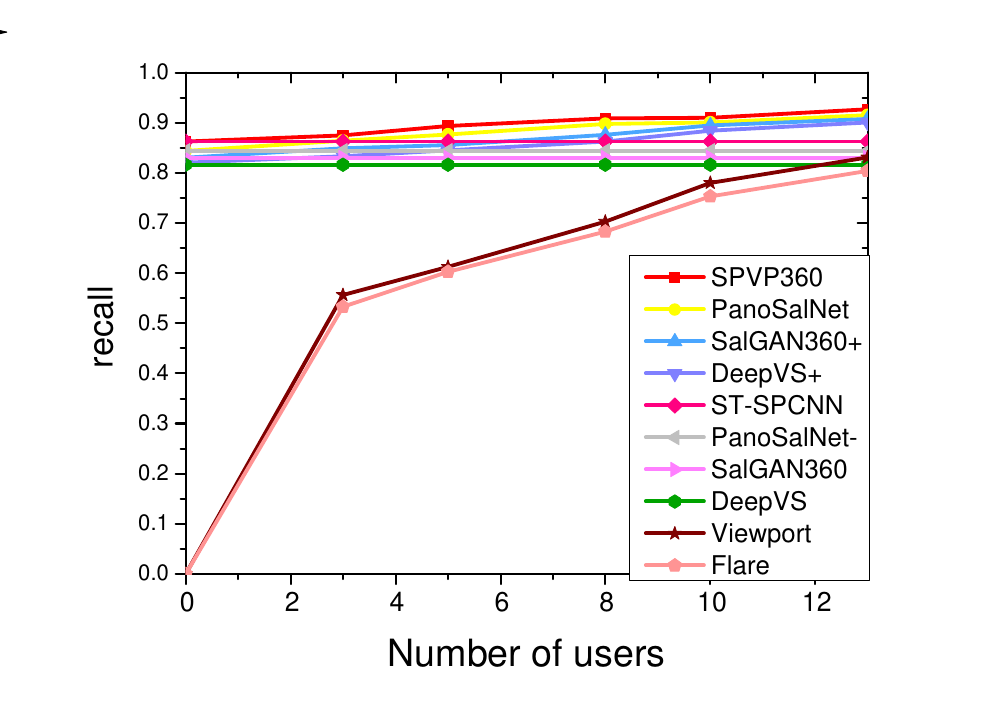}
 \end{minipage}
\label{fig_vp_num_re}
}
\vspace{-0.1in}
\caption{The effect of the number of users sending FoV on the prediction performance.}
\label{fig_vp_num_all}
\vspace{-0.1in}
\end{figure}

\begin{figure}[htb]
\centering
\vspace{-0.3in}
\subfigure[]{
   \begin{minipage}[t]{1\linewidth}
			\centering
			\vspace{0pt}
			\includegraphics[width=1\textwidth]{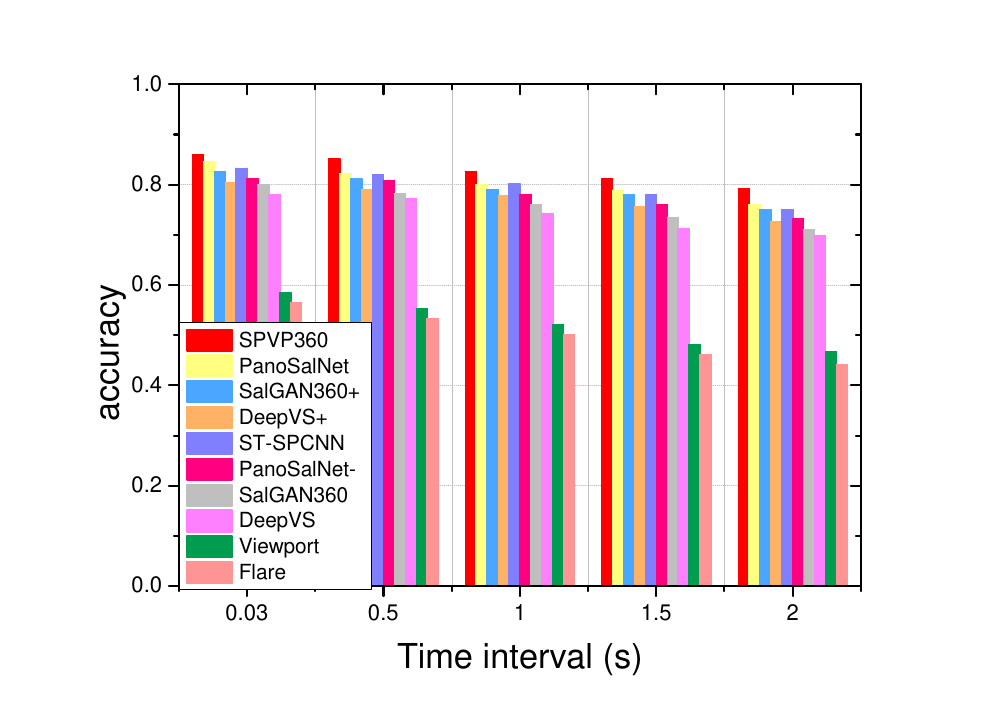}
    \end{minipage}
\label{fig_vp_ac:subfig_a}
}

\vspace{-0.3in}
\subfigure[]{
 \begin{minipage}[t]{1\linewidth}
			\centering
			\vspace{0pt}
			\includegraphics[width=1\textwidth]{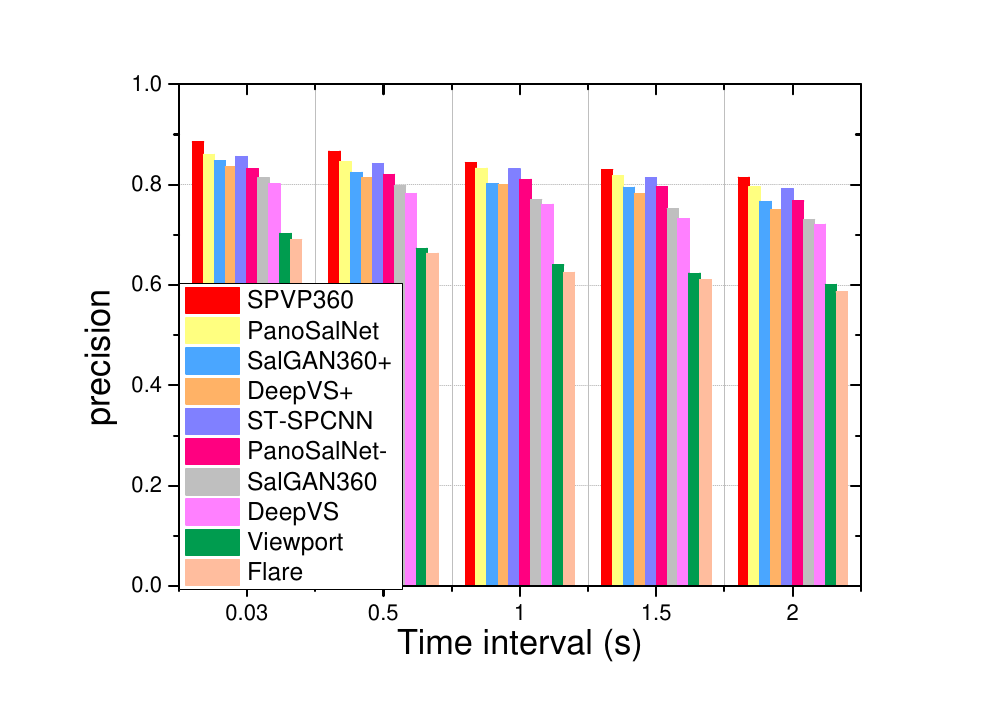}
 \end{minipage}
\label{fig_vp_pe:subfig_b}
}

\vspace{-0.3in}
\subfigure[]{
 \begin{minipage}[t]{1\linewidth}
			\centering
			\vspace{0pt}
			\includegraphics[width=1\textwidth]{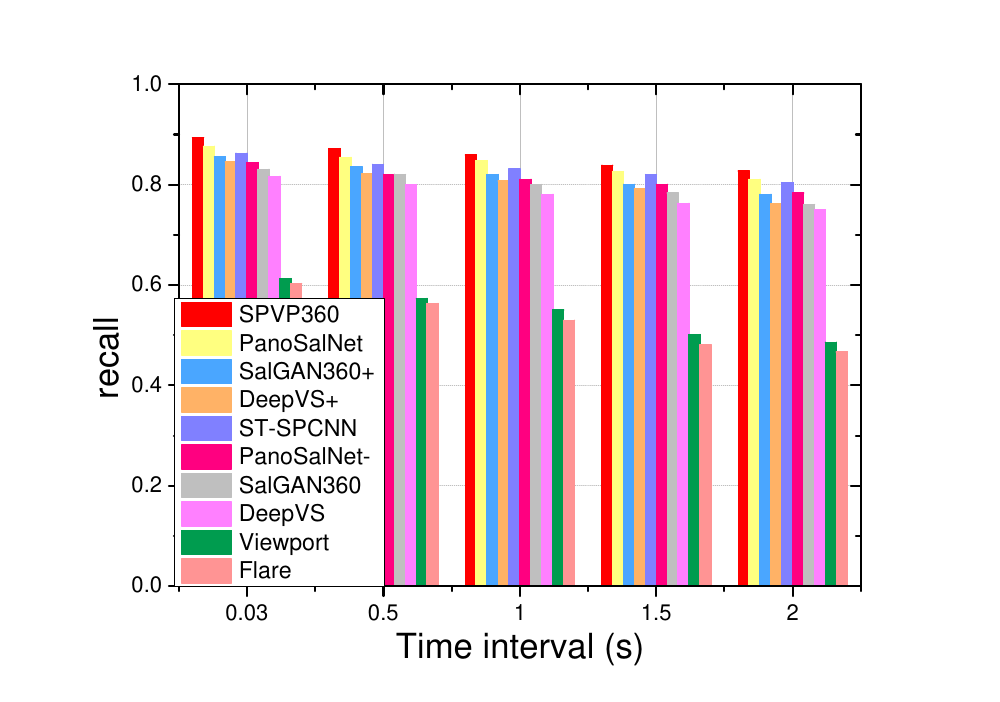}
 \end{minipage}
\label{fig_vp_re:subfig_c}
}

\caption{Impact of $\varkappa$ on prediction accuracy, precision and recall.}
\label{fig_vp_in_all}
\vspace{-0.1in}
\end{figure}

\textbf{FoV prediction results:} During the experiments, different time intervals and different numbers of users sending FoV feedbacks are set for FoV prediction; the results are shown in Fig. \ref{fig_vp_vall}-\ref{fig_vp_in_all}. The relationship between the number of feedback users (i.e., users sending FoV) and the accuracy and latency as well as the computational effort is also studied. The delay time indicates the time required for one prediction and the computational effort is determined using FLOPs. In addition, the effect of different head movement frequencies on FoV prediction is also investigated.

As illustrated in Fig. \ref{fig_vp_vall}, five users' FoV feedback is randomly selected, and the prediction interval $\varkappa=0.03s$ is used for prediction. The first four sets of results are averages of FoV predictions for four individual videos, respectively. The last set of data is the average of FoV predictions for all 360-degree videos, with a total of 10 videos. From the experimental results, we can see that our prediction method outperforms other methods in terms of accuracy, precision and recall. Although the prediction performance of our model degrades in some videos, our method still performs better than the comparison methods. 
In addition, we set the confidence level to $95\%$ with a confidence interval of $0.8589 \pm  0.088$ for accuracy, $0.8856 \pm 0.0715$ for precision, and $0.8933 \pm 0.0652$ for recall.

The next step is to investigate the effect of the number of users sending FoV on the prediction results. Different numbers of feedback users are selected, and $\varkappa_{0}=0.03s$ is used for prediction.
When there is no feedback user, only video saliency information is used for FoV prediction, and the results are shown in Fig. \ref{fig_vp_num_all}. The FoV prediction methods combining saliency detection and users' feedback FoV have higher-accuracy results than the prediction methods considering only video saliency. Moreover, the accuracy becomes better as the number of feedback users increases. Furthermore, Viewport outperforms even DeepVS and SalGAN360 as the number of user FoVs increases. 

\begin{table*}[ht]
    \vspace{-0.1in}
	\centering
	\caption{The impact of the number of users sending FoV.} 
	\label{table_tradeoff}
	\centering
	\renewcommand\arraystretch{1.35} 
    \setlength{\tabcolsep}{5mm}{
	\begin{tabular}{|c|c|c|c|}
    \hline
	The number of users &  Accuracy  & Latency($s$) & Computational requirement (GFLOPs) \\ \hline
    2          & 0.8463     &  0.2732       &       5.2817      \\ \hline
    5          & 0.8589     &  0.3101      &       7.3298        \\ \hline
    8          & 0.8709     &  0.3428       &       9.3779    \\ \hline
    11         & 0.8941     &  0.3874      &       11.426     \\ \hline
    \end{tabular}}
\end{table*}

In order to further study how the number of feedback users affect the FoV prediction, different number of feedback users are selected and the prediction interval is set to $\varkappa_{0}=0.03 $ on the Nvidia Titan 2060i GPU for prediction. The relationship between different numbers of feedback users, prediction accuracy, latency, and computational requirement is shown in Table \ref{table_tradeoff}. From this table, we can see that the accuracy, latency, and computational requirement  increase with the increase of the number of feedback users. Higher accuracy, lower latency and lower computational requirement indicate better performance of the prediction method. When more than 10 users are selected for prediction, the computational power of the Nvidia Titan 2060i GPU is capped and latency is greatly increased. Note that considering that latency is important for prediction, we cannot choose too many users and must make a trade-off between system requirements and performance.

\begin{table}[ht]
	\centering
	\caption{The time required for prediction on different devices.}
	\label{tab_GPU}
	\centering
	\renewcommand\arraystretch{1.35} 
    \setlength{\tabcolsep}{2mm}{
	\begin{tabular}{|c|c|c|}
    \hline
	The GPU type            & SP-STCNN(s) & SP-ConvGRU(s) \\\hline
    Nvidia Tesla V100   &	0.1253    &	0.0235        \\ \hline
    Nvidia Titan 2060i  &	0.2528    &	0.0573      \\ \hline
    \end{tabular}
    }
\end{table}

We have also tested it on Nvidia Tesla V100 GPU and Nvidia Titan 2060i GPU, respectively. The results are shown in Table \ref{tab_GPU}. The prediction times are 0.1488s and 0.3101s on Nvidia Tesla V10 GPU and Nvidia Titan 2060i GPU, respectively. Note that this is obtained without code optimization and we believe this is fast enough for a seamless video experience. In addition, the performance of the mobile GPU is similar to that of the Nvidia Titan 2060i GPU \cite{liang2020ai}, which proves the feasibility of the proposed solution in mobile devices.

\begin{table}[ht]
	\centering
	\caption{The effects of different head movement frequency on prediction performance.}
	\label{tab_fre_type}
	\centering
	\renewcommand\arraystretch{1.35} 
    \setlength{\tabcolsep}{5mm}{
	\begin{tabular}{|c|c|c|c|}
    \hline
	Type    &  Accuracy & Precision  & Recall     \\ \hline
    More    &  0.8545   &  0.8781    & 0.8872     \\ \hline
    Middle  &  0.8589   &  0.8856    & 0.8933     \\ \hline
    Less    &  0.8601   &  0.8912    & 0.9001     \\ \hline
    \end{tabular}}
\end{table}	

We next investigate the effect of users with different head movement frequencies on the prediction results. We select five users from each of the different head motion frequency types (i.e., More, Middle, Less) for FoV prediction. The prediction interval is set to $\varkappa_{0}=0.03s $.
The performance of our prediction method for different head movement types is shown in Table \ref{tab_fre_type}. We can see that there is little difference in the prediction performance of our method in these three cases, although our method performs only slightly better when the head motion frequency is low, which demonstrates the feasibility of the proposed scheme in the presence of different user head movements.

Next, the effect of time interval on FoV prediction is evaluated. Five feedback users' FoVs are selected; the results are shown in Fig. \ref{fig_vp_in_all}. We can see  that the prediction accuracy of our method decreases slightly as the time interval increases, however, the performance of our method is still better than the comparison methods, demonstrating the effectiveness of the proposed scheme.

\begin{figure*}[htb!]
    \vspace{-0.1in}
    \centering
    \includegraphics[width=1\textwidth]{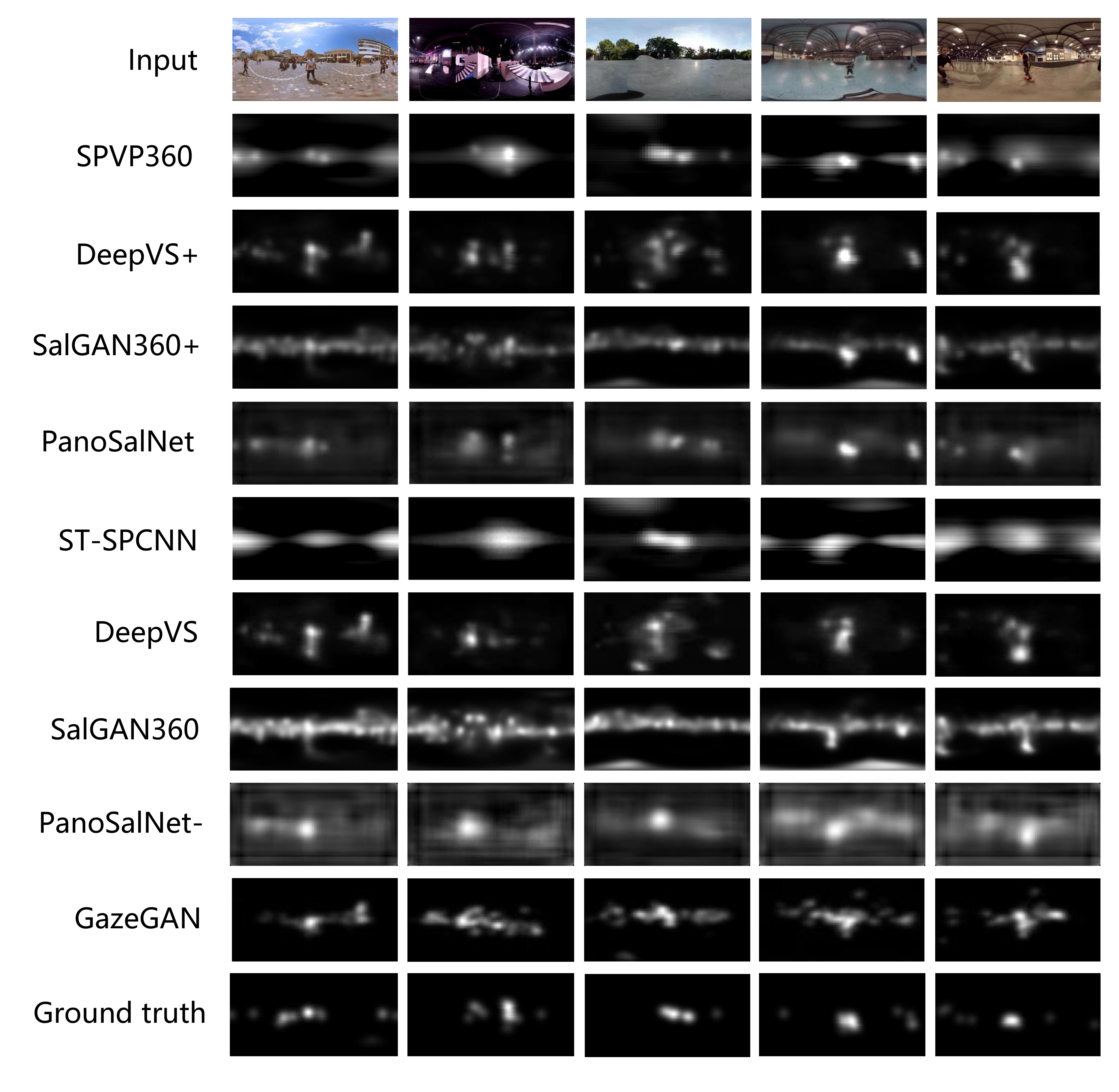}
    \vspace{-0.3in}
    \caption{ Saliency maps of five randomly selected videos generated by the proposed method and by seven other methods, as well as the ground truth.}
    \label{fig_spc_vpsal}
\end{figure*}

In addition, the results of different prediction methods are visualized, as shown in Fig. \ref{fig_spc_vpsal}. It is apparent that SalGAN360+ and PanoSalNet- incorrectly treat almost all equatorial regions as salient regions, whereas DeepVS ignores many salient regions. ST-SPCNN extracts a saliency map that is close to the ground truth. However, this does not mean that the ST-SPCNN saliency regions are error-free. With the addition of Viewport in SPVP360, SalGAN360+, and PanoSalNet, the regions that should not have been considered significant are corrected to some extent. The significant regions ignored by DeepVS+ are also corrected to some extent by adding FoV. However, the proposed model's FoV prediction map is still closer to the ground truth than the others. In summary, both qualitative and quantitative results show that the prediction forecast from the proposed model is the most effective.

\begin{figure*}[htb]
\centering
\vspace{-0.4in}
\subfigure[The NSS results in dataset1.]{
   \begin{minipage}[b]{0.48\linewidth}
	\centering
	\vspace{0pt}
	\includegraphics[width=1\textwidth]{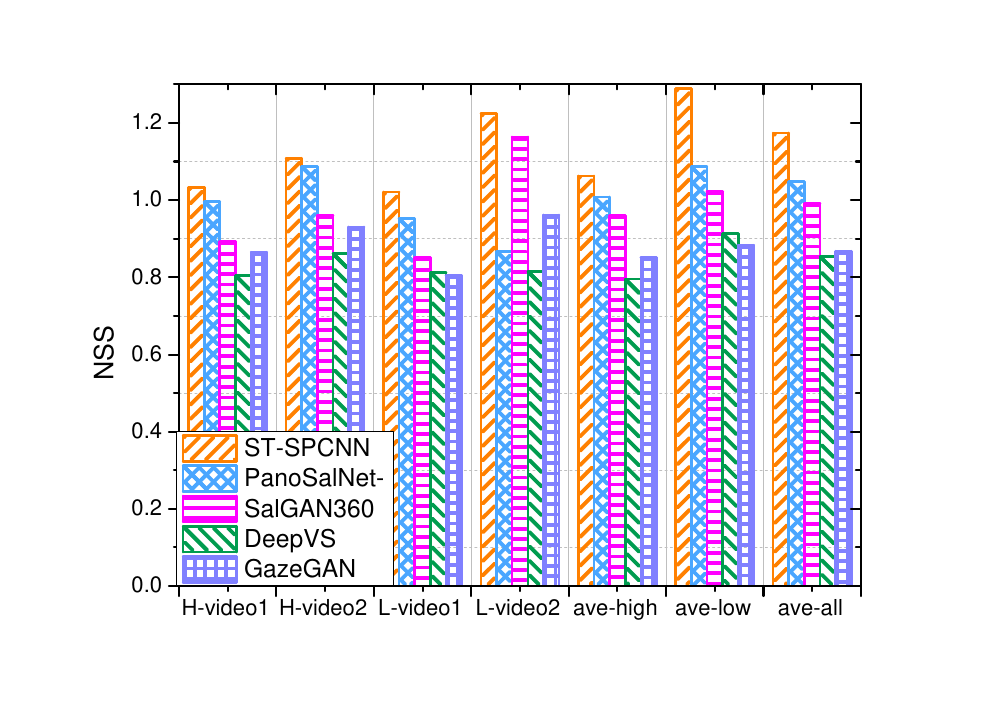}
    \end{minipage}%
\label{fig_sal_nss:subfig_a1}
}
\subfigure[The NSS results in dataset2.]{
 \begin{minipage}[b]{0.48\textwidth}
		\centering
		\vspace{0pt}
		\includegraphics[width=1\textwidth]{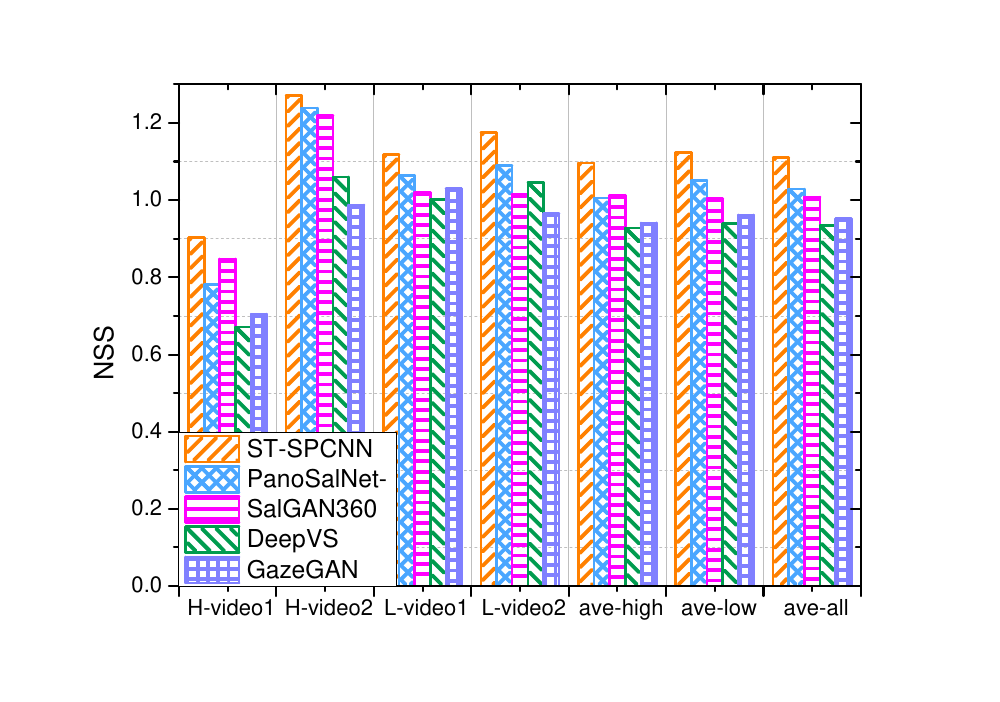}%
 \end{minipage}
\label{fig_sal_nss:subfig_a2}
}

\vspace{-0.3in}
\subfigure[The CC results in dataset1.]{
 \begin{minipage}[t]{0.48\linewidth}
			\centering
			\vspace{0pt}
			\includegraphics[width=1\textwidth]{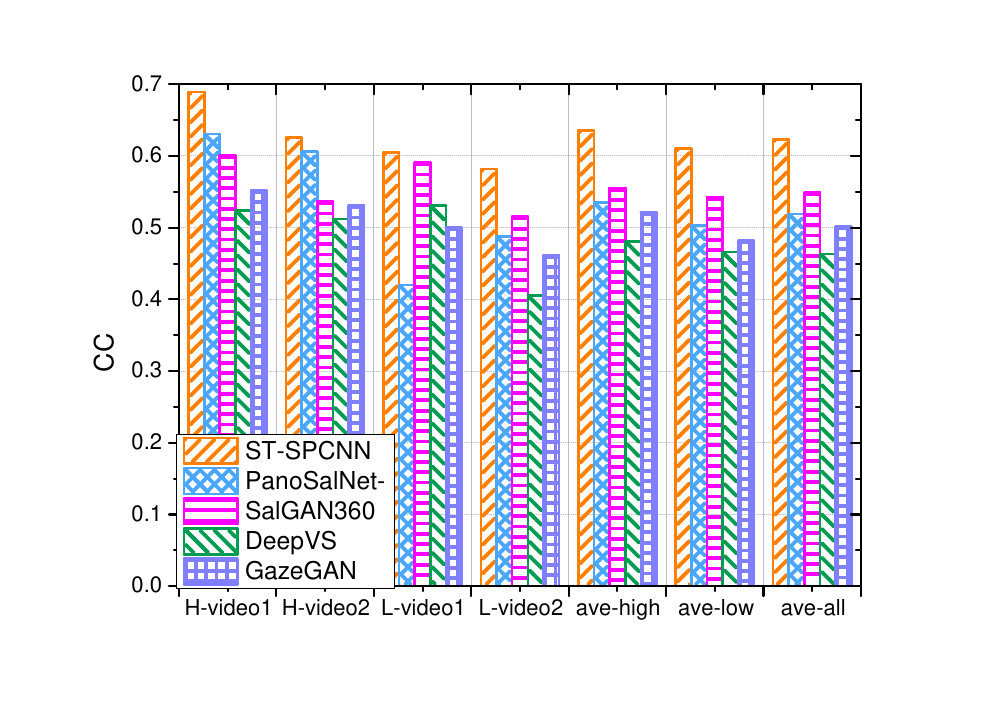}
 \end{minipage}%
\label{fig_sal_cc:subfig_b1}
}
\subfigure[The CC results in dataset2.]{
 \begin{minipage}[t]{0.48\linewidth}
			\centering
			\vspace{0pt}
			\includegraphics[width=1\textwidth]{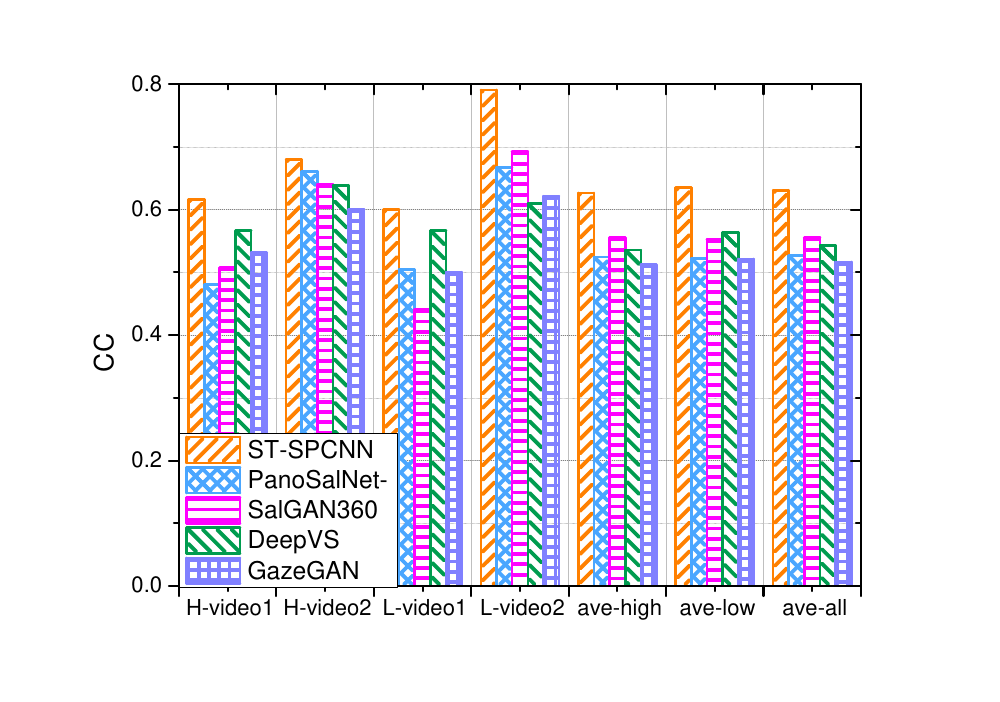}
 \end{minipage}
\label{fig_sal_cc:subfig_b2}
}

\vspace{-0.3in}   
\subfigure[The AUC results in dataset1.]{
 \begin{minipage}[t]{0.48\linewidth}
			\centering
			\vspace{0pt}
			\includegraphics[width=1\textwidth]{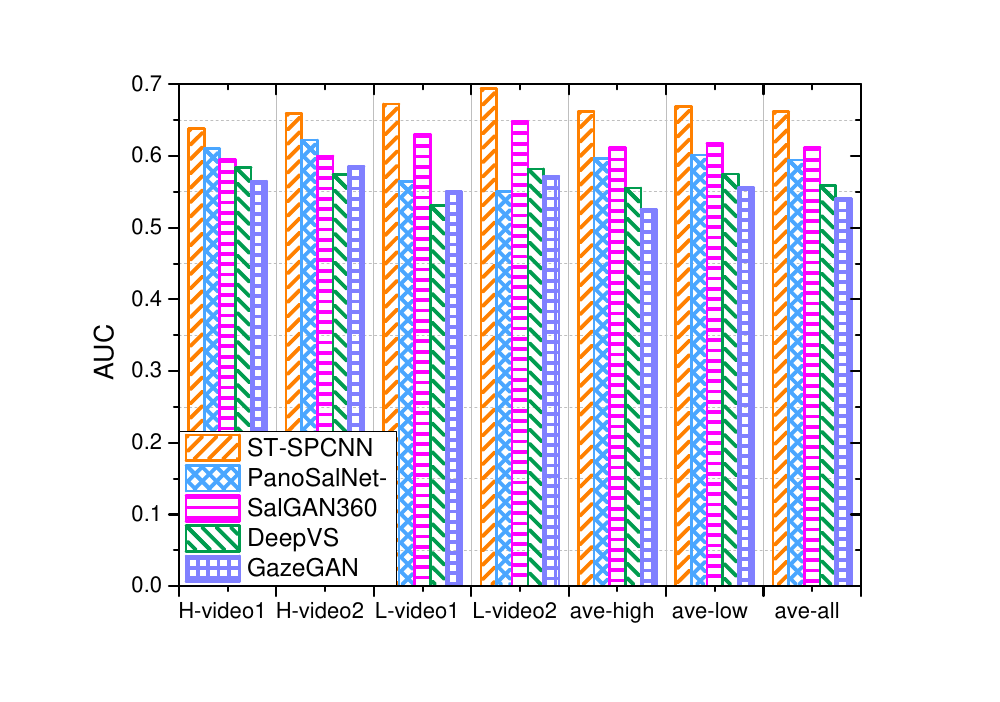}
 \end{minipage}
\label{fig_sal_auc:subfig_c1}
   }
\subfigure[The AUC results in dataset2.]{
\begin{minipage}[t]{0.48\linewidth}
			\centering
			\vspace{0pt}
			\includegraphics[width=1\textwidth]{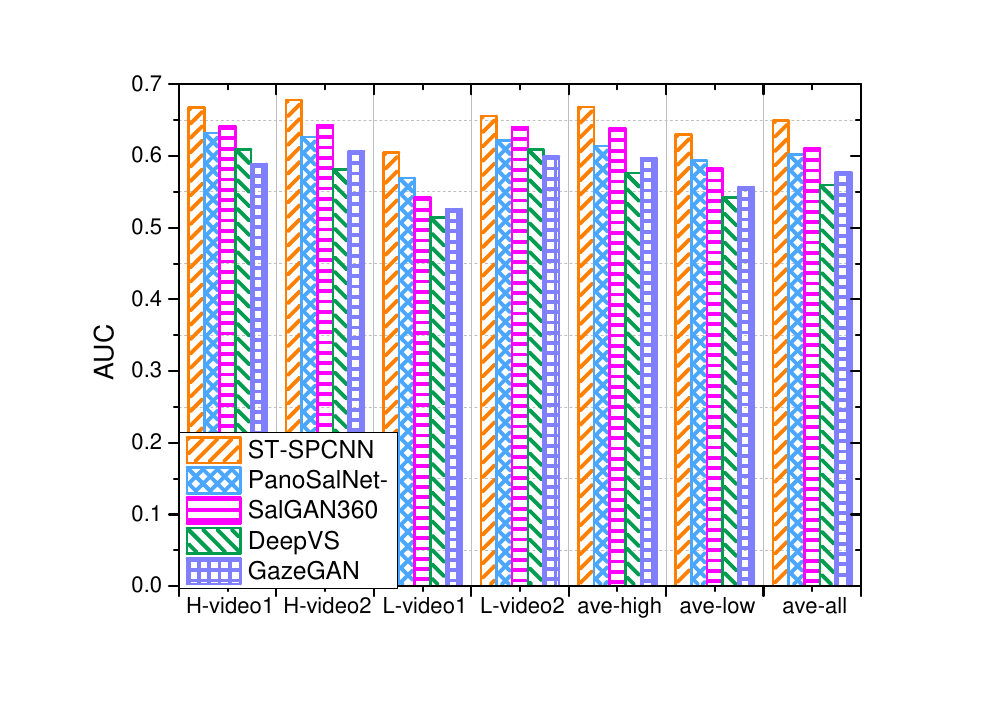}
		\end{minipage}
\label{fig_sal_auc:subfig_c2}
}

\vspace{-0.1in}
\caption{The performance comparison for saliency detection.}    
\label{fig:compare_fig}
\end{figure*}

\textbf{Saliency detection results:} The performance of the proposed saliency detection model is also evaluated. The results of ST-SPCNN are compared with PanoSalNet-, SalGAN360, DeepVS and GazeGAN. The NSS, CC, and AUC results are shown in Fig. \ref{fig:compare_fig}. The first four sets of results are averaged over the saliency maps of four individual videos, respectively. The middle two sets of results are averaged over the saliency maps of the high-motion videos and the low-motion videos, respectively. The last set of data is averaged over the saliency map of all the 360-degree videos, representing a total of ten videos. Higher values of NSS, CC, and AUC imply more accurate saliency detection. From the Fig \ref{fig_sal_nss:subfig_a1}-\ref{fig_sal_cc:subfig_b1}-\ref{fig_sal_auc:subfig_c1}, we can see that the proposed saliency detection method significantly outperforms the other methods on three evaluation metrics, which proves the effectiveness of the proposed 360-degree video saliency prediction model. The proposed ST-SPCNN eliminates the effect of distortion by convolution, which greatly improves its performance. Although SalGAN360 proposes changes to the projection to minimize the effects of distortion, its performance is still limited. This may have been the case because each individual face output has its own unique saliency region, and combining the individual face outputs produces many more insignificant regions in the saliency detection results. PanoSalNet- is designed for 360-degree videos, DeepVS is designed for 2D videos fixed in a single FoV, and GazeGAN considers data enhancement strategies to enable better performance of saliency detection in 2D videos. However, they do not consider the effect of video distortion, resulting in poor performance of saliency detection in 360 videos.


We have also evaluated our model on dataset2 \cite{8578657} and  selected eight videos for testing. In Fig \ref{fig_sal_nss:subfig_a2}-\ref{fig_sal_cc:subfig_b2}-\ref{fig_sal_auc:subfig_c2}, the first four sets of results are averaged over the saliency maps of the individual videos, respectively. The middle two sets of results are averaged over the saliency maps of the high motion videos and the low motion videos, respectively, and the last set of data is averaged over the saliency maps of all 360 degree videos. Our proposed saliency detection method continues to outperform the other four methods on all three metrics, which proves that the proposed 360-degree video saliency prediction model is effective.

We next evaluate the contribution of the different components of the model in the saliency detection, and the ablation experiments are performed using the first dataset \cite{Zhang_2018_ECCV}. For this purpose, we have constructed four different variants, including: (1) S-SPCNN; (2) T-SPCNN; (3) ST-SPCNN- (Indicates no attention mechanism); and (4) ST-SPCNN. And we also compare the variants set in our experiments with the existing method ST-CNN \cite{Jiang_2018_ECCV}.

\begin{table}[ht]
	\centering
	\caption{The performance of  different network components.}
	\centering
	\renewcommand\arraystretch{1.35} 
    \setlength{\tabcolsep}{5mm}{
	\begin{tabular}{|c|c|c|c|c|}
    \hline
	Components & NSS      & CC        & AUC       \\    \hline
    S-SPCNN    &  0.6301  & 0.4058  &  0.5471 \\ \hline
    S-CNN      & 0.5819   & 0.3837  &  0.5173 \\ \hline
    T-SPCNN    & 0.782    & 0.4384   &  0.5762 \\ \hline
    T-CNN      & 0.617    & 0.4137   &  0.5335 \\ \hline
    ST-SPCNN-  & 0.9653   & 0.5587   &  0.6069 \\ \hline
    ST-CNN     & 0.8487   & 0.4634   &  0.5589 \\ \hline
    \end{tabular}}
     \vspace{-0.1in}
     \label{tab_dif_comp}
\end{table}

\textbf{The impact of different components.} We compare the performance of the spherical convolution-based network with existing networks. S-SPCNN and S-CNN denote video spatial feature extraction networks. T-SPCNN and T-CNN denote video temporal feature extraction networks. ST-SPCNN and ST-CNN denote spatial-temporal feature extraction networks. S-SPCNN, T-SPCNN and ST-SPCNN- denote the spherical convolution-based networks. From Table \ref{tab_dif_comp},
firstly, the performance of the combined spatial-temporal feature network outperforms that of the spatial feature extraction network alone and the temporal feature extraction network alone, indicating that considering both temporal and spatial features of the video can yield better results. Secondly, the performance of the spherical convolution-based network is higher than the existing methods. This is because the distortion of the 2D plane of the 360-degree video projection causes the parameter sharing of the 2D convolution to fail, which makes the network detection less effective. This also indicates that spherical convolution can eliminate the effect of distortion and thus improve the performance of the saliency detection model.

\begin{table}[ht]
	\centering
	\caption{The impact of different CBAM.}
	\vspace{-0.05in}
	\label{tab_cbam}
	\centering
	\renewcommand\arraystretch{1.35} 
    \setlength{\tabcolsep}{3mm}{
	\begin{tabular}{|c|c|c|c|c|}
    \hline
	CBAM            & NSS       & CC       & AUC     \\\hline
    no CBAM        & 0.9653    & 0.5587    & 0.6069   \\ \hline
    with 2d-CBAM   & 1.0271    & 0.5765   & 0.6278   \\ \hline
    with SP-CBAM   & 1.1736    & 0.6234   & 0.6619  \\ \hline
    \end{tabular}}
\end{table}

\textbf{The impact of CBAM.} We have also tested the performance of the attention module with different convolutions. Table \ref{tab_cbam} shows the possible configurations, where the first one indicates the inclusion of no attention mechanism, the second one indicates the inclusion of an attention mechanism based on 2D convolution, and the last one indicates the inclusion of an attention mechanism based on spherical convolution. Clearly, the results in the table indicate that adding the attention module based on spherical convolution yields better saliency detection results.

\section{Conclusions}\label{sec_conclusion}
This paper has proposed a spherical convolution-based method, SPVP360, which is a multi-source FoV prediction method combining salient features extracted from 360-degree video with limited FoV feedback information. A spherical convolutional network was used instead of a traditional 2D convolutional network to eliminate the problem of weight sharing failure in the convolutional network caused by video projection distortion. Specifically, salient spatial-temporal features were extracted through a spherical convolution-based saliency detection model, ST-SPCNN, a small amount of user-feedback FoV information was represented as a time-series model based on the spherical convolution-based GRU network, and finally the extracted salient video features were combined to predict future user FoVs. The approach taken by the spherical convolution-based saliency detection model (ST-SPCNN) is to first extract spatial-temporal features using S-SPCNN and T-SPCNN, and then to actively enhance spatial-temporal features by the attention mechanism and map them into the saliency map. The proposed method is compared with several FoV prediction methods on two 360-degree video dataset. The experimental results show that the performance of the proposed method is better than other prediction methods. 

\section{Acknowledgments}
This work is supported in part by grants from the National Natural Science Foundation of China (52077049), Anhui Provincial Natural Science Foundation (2008085UD04), and Fundamental Research Funds for the Central Universities (PA2020GDJQ0027).

\bibliographystyle{IEEEtran} 
\bibliography{ref}

\begin{thebibliography}{10}
\providecommand{\url}[1]{#1}
\csname url@samestyle\endcsname
\providecommand{\newblock}{\relax}
\providecommand{\bibinfo}[2]{#2}
\providecommand{\BIBentrySTDinterwordspacing}{\spaceskip=0pt\relax}
\providecommand{\BIBentryALTinterwordstretchfactor}{4}
\providecommand{\BIBentryALTinterwordspacing}{\spaceskip=\fontdimen2\font plus
\BIBentryALTinterwordstretchfactor\fontdimen3\font minus
  \fontdimen4\font\relax}
\providecommand{\BIBforeignlanguage}[2]{{%
\expandafter\ifx\csname l@#1\endcsname\relax
\typeout{** WARNING: IEEEtran.bst: No hyphenation pattern has been}%
\typeout{** loaded for the language `#1'. Using the pattern for}%
\typeout{** the default language instead.}%
\else
\language=\csname l@#1\endcsname
\fi
#2}}
\providecommand{\BIBdecl}{\relax}
\BIBdecl

\bibitem{liu2021point}
Z.~Liu, Q.~Li, X.~Chen, C.~Wu, S.~Ishihara, J.~Li, and Y.~Ji, ``Point cloud
  video streaming: Challenges and solutions,'' \emph{IEEE Network}, vol.~35,
  no.~5, pp. 202--209, 2021.

\bibitem{fan2019survey}
C.-L. Fan, W.-C. Lo, Y.-T. Pai, and C.-H. Hsu, ``A survey on 360 video
  streaming: Acquisition, transmission, and display,'' \emph{ACM Computing
  Surveys (CSUR)}, vol.~52, no.~4, pp. 1--36, 2019.

\bibitem{8795062}
A.~Xu, X.~Chen, Y.~Liu, and Y.~Wang, ``A flexible viewport-adaptive processing
  mechanism for real-time vr video transmission,'' in \emph{2019 IEEE
  International Conference on Multimedia Expo Workshops (ICMEW)}, 2019, pp.
  336--341.

\bibitem{zhang2020360}
J.~Zhang, Y.~Zhong, Y.~Han, D.~Li, C.~Yu, and J.~Mo, ``A 360 \degree video
  adaptive streaming scheme based on multiple video qualities,'' in \emph{2020
  IEEE/ACM 13th International Conference on Utility and Cloud Computing
  (UCC)}.\hskip 1em plus 0.5em minus 0.4em\relax IEEE, 2020, pp. 402--407.

\bibitem{8957509}
J.~Li, R.~Feng, W.~Sun, Z.~Liu, and Q.~Li, ``Qoe-driven coupled uplink and
  downlink rate adaptation for 360-degree video live streaming,'' \emph{IEEE
  Communications Letters}, vol.~24, no.~4, pp. 863--867, 2020.

\bibitem{guo2018optimal}
C.~Guo, Y.~Cui, and Z.~Liu, ``Optimal multicast of tiled 360 vr video,''
  \emph{IEEE Wireless Communications Letters}, vol.~8, no.~1, pp. 145--148,
  2018.

\bibitem{guo2021power}
C.~Guo, L.~Zhao, Y.~Cui, Z.~Liu, and D.~W.~K. Ng, ``Power-efficient wireless
  streaming of multi-quality tiled 360 vr video in mimo-ofdma systems,''
  \emph{IEEE Transactions on Wireless Communications}, vol.~20, no.~8, pp.
  5408--5422, 2021.

\bibitem{8999740}
O.~Eltobgy, O.~Arafa, and M.~Hefeeda, ``Mobile streaming of live 360-degree
  videos,'' \emph{IEEE Transactions on Multimedia}, vol.~22, no.~12, pp.
  3139--3152, 2020.

\bibitem{zhai2020perceptual}
G.~Zhai and X.~Min, ``Perceptual image quality assessment: a survey,''
  \emph{Science China Information Sciences}, vol.~63, no.~11, pp. 211--301,
  2020.

\bibitem{8281422}
K.~Lee, G.~Guerrero, S.~Cha, Y.~Kim, and S.~Cho, ``Vr theater, a virtual
  reality based multi-screen movie theater simulator for verifying multi-screen
  content and environment,'' in \emph{SMPTE 2017 Annual Technical Conference
  and Exhibition}.\hskip 1em plus 0.5em minus 0.4em\relax SMPTE, 2017, pp.
  1--13.

\bibitem{rumney2013lte}
M.~Rumney \emph{et~al.}, \emph{LTE and the evolution to 4G wireless: Design and
  measurement challenges}.\hskip 1em plus 0.5em minus 0.4em\relax John Wiley \&
  Sons, 2013.

\bibitem{7840720}
Y.~Bao, H.~Wu, T.~Zhang, A.~A. Ramli, and X.~Liu, ``Shooting a moving target:
  Motion-prediction-based transmission for 360-degree videos,'' in \emph{2016
  IEEE International Conference on Big Data (Big Data)}.\hskip 1em plus 0.5em
  minus 0.4em\relax IEEE, 2016, pp. 1161--1170.

\bibitem{nasrabadi2020viewport}
A.~T. Nasrabadi, A.~Samiei, and R.~Prakash, ``Viewport prediction for 360
  \degree videos: A clustering approach,'' in \emph{Proceedings of the 30th ACM
  Workshop on Network and Operating Systems Support for Digital Audio and
  Video}, ser. NOSSDAV '20.\hskip 1em plus 0.5em minus 0.4em\relax New York,
  NY, USA: Association for Computing Machinery, 2020, p. 34–39.

\bibitem{9207562}
J.~Tang, Y.~Huo, S.~Yang, and J.~Jiang, ``A viewport prediction framework for
  panoramic videos,'' in \emph{2020 International Joint Conference on Neural
  Networks (IJCNN)}.\hskip 1em plus 0.5em minus 0.4em\relax IEEE, 2020, pp.
  1--8.

\bibitem{8613652}
S.~Petrangeli, G.~Simon, and V.~Swaminathan, ``Trajectory-based viewport
  prediction for 360-degree virtual reality videos,'' in \emph{2018 IEEE
  International Conference on Artificial Intelligence and Virtual Reality
  (AIVR)}, 2018, pp. 157--160.

\bibitem{9238515}
J.~Chen, X.~Luo, M.~Hu, D.~Wu, and Y.~Zhou, ``Sparkle: User-aware viewport
  prediction in 360-degree video streaming,'' \emph{IEEE Transactions on
  Multimedia}, vol.~23, pp. 3853--3866, 2021.

\bibitem{nguyen2018your}
A.~Nguyen, Z.~Yan, and K.~Nahrstedt, ``Your attention is unique: Detecting
  360-degree video saliency in head-mounted display for head movement
  prediction,'' in \emph{Proceedings of the 26th ACM International Conference
  on Multimedia}, ser. MM '18.\hskip 1em plus 0.5em minus 0.4em\relax New York,
  NY, USA: Association for Computing Machinery, 2018, p. 1190–1198.

\bibitem{8551543}
F.-Y. Chao, L.~Zhang, W.~Hamidouche, and O.~Deforges, ``Salgan360: Visual
  saliency prediction on 360 degree images with generative adversarial
  networks,'' in \emph{2018 IEEE International Conference on Multimedia Expo
  Workshops (ICMEW)}.\hskip 1em plus 0.5em minus 0.4em\relax IEEE, 2018, pp.
  01--04.

\bibitem{2017Flat2Sphere}
Y.-C. Su and K.~Grauman, ``Learning spherical convolution for fast features
  from 360 \degree imagery,'' in \emph{Proceedings of the 31st International
  Conference on Neural Information Processing Systems}, ser. NIPS'17.\hskip 1em
  plus 0.5em minus 0.4em\relax Curran Associates, Inc., 2017, p. 529–539.

\bibitem{8971933}
X.~Li, S.~Wang, C.~Zhu, L.~Song, R.~Xie, and W.~Zhang, ``Viewport prediction
  for panoramic video with multi-cnn,'' in \emph{2019 IEEE International
  Symposium on Broadband Multimedia Systems and Broadcasting (BMSB)}.\hskip 1em
  plus 0.5em minus 0.4em\relax IEEE, 2019, pp. 1--6.

\bibitem{yu2015framework}
M.~Yu, H.~Lakshman, and B.~Girod, ``A framework to evaluate omnidirectional
  video coding schemes,'' in \emph{2015 IEEE International Symposium on Mixed
  and Augmented Reality}.\hskip 1em plus 0.5em minus 0.4em\relax IEEE, 2015,
  pp. 31--36.

\bibitem{8702654}
Q.~Yang, J.~Zou, K.~Tang, C.~Li, and H.~Xiong, ``Single and sequential
  viewports prediction for 360-degree video streaming,'' in \emph{2019 IEEE
  International Symposium on Circuits and Systems (ISCAS)}, 2019, pp. 1--5.

\bibitem{8942328}
J.~Vielhaben, H.~Camalan, W.~Samek, and M.~Wenzel, ``Viewport forecasting in
  360 \degree virtual reality videos with machine learning,'' in \emph{2019
  IEEE International Conference on Artificial Intelligence and Virtual Reality
  (AIVR)}.\hskip 1em plus 0.5em minus 0.4em\relax IEEE, 2019, pp. 74--747.

\bibitem{9180528}
M.~Jamali, S.~Coulombe, A.~Vakili, and C.~Vazquez, ``Lstm-based viewpoint
  prediction for multi-quality tiled video coding in virtual reality
  streaming,'' in \emph{2020 IEEE International Symposium on Circuits and
  Systems (ISCAS)}.\hskip 1em plus 0.5em minus 0.4em\relax IEEE, 2020, pp.
  1--5.

\bibitem{2021PARIMA}
L.~Chopra, S.~Chakraborty, A.~Mondal, and S.~Chakraborty, ``Parima: Viewport
  adaptive 360-degree video streaming,'' in \emph{Proceedings of the Web
  Conference 2021}, ser. WWW '21.\hskip 1em plus 0.5em minus 0.4em\relax New
  York, NY, USA: Association for Computing Machinery, 2021, p. 2379–2391.

\bibitem{2020Flocking}
L.~Sun, Y.~Mao, T.~Zong, Y.~Liu, and Y.~Wang, ``Flocking-based live streaming
  of 360-degree video,'' in \emph{Proceedings of the 11th ACM Multimedia
  Systems Conference}, ser. MMSys '20.\hskip 1em plus 0.5em minus 0.4em\relax
  New York, NY, USA: Association for Computing Machinery, 2020, p. 26–37.

\bibitem{yang2020predicting}
Z.~Yang, L.~Huang, Y.~Chen, Z.~Wei, S.~Ahn, G.~Zelinsky, D.~Samaras, and
  M.~Hoai, ``Predicting goal-directed human attention using inverse
  reinforcement learning,'' in \emph{Proceedings of the IEEE/CVF conference on
  computer vision and pattern recognition}, 2020, pp. 193--202.

\bibitem{8578657}
Y.~Xu, Y.~Dong, J.~Wu, Z.~Sun, Z.~Shi, J.~Yu, and S.~Gao, ``Gaze prediction in
  dynamic 360 \degree immersive videos,'' in \emph{2018 IEEE/CVF Conference on
  Computer Vision and Pattern Recognition}, 2018, pp. 5333--5342.

\bibitem{9072511}
M.~Qiao, M.~Xu, Z.~Wang, and A.~Borji, ``Viewport-dependent saliency prediction
  in 360 \degree video,'' \emph{IEEE Transactions on Multimedia}, vol.~23, pp.
  748--760, 2021.

\bibitem{9089486}
X.~Feng, Y.~Liu, and S.~Wei, ``Livedeep: Online viewport prediction for live
  virtual reality streaming using lifelong deep learning,'' in \emph{2020 IEEE
  Conference on Virtual Reality and 3D User Interfaces (VR)}.\hskip 1em plus
  0.5em minus 0.4em\relax IEEE, 2020, pp. 800--808.

\bibitem{9018228}
X.~Chen, A.~T.~Z. Kasgari, and W.~Saad, ``Deep learning for content-based
  personalized viewport prediction of 360-degree vr videos,'' \emph{IEEE
  Networking Letters}, vol.~2, no.~2, pp. 81--84, 2020.

\bibitem{8683125}
B.~Dedhia, J.-C. Chiang, and Y.-F. Char, ``Saliency prediction for
  omnidirectional images considering optimization on sphere domain,'' in
  \emph{ICASSP 2019 - 2019 IEEE International Conference on Acoustics, Speech
  and Signal Processing (ICASSP)}.\hskip 1em plus 0.5em minus 0.4em\relax IEEE,
  2019, pp. 2142--2146.

\bibitem{8911454}
Y.~Zhang, F.~Dai, Y.~Ma, H.~Li, Q.~Zhao, and Y.~Zhang, ``Saliency prediction
  network for $360^\circ$ videos,'' \emph{IEEE Journal of Selected Topics in
  Signal Processing}, vol.~14, no.~1, pp. 27--37, 2020.

\bibitem{zhao2019laddernet}
P.~Zhao, Y.~Zhang, K.~Bian, H.~Tuo, and L.~Song, ``Laddernet: Knowledge
  transfer based viewpoint prediction in 360 \degree video,'' in \emph{ICASSP
  2019-2019 IEEE International Conference on Acoustics, Speech and Signal
  Processing (ICASSP)}.\hskip 1em plus 0.5em minus 0.4em\relax IEEE, 2019, pp.
  1657--1661.

\bibitem{8779683}
C.-L. Fan, S.-C. Yen, C.-Y. Huang, and C.-H. Hsu, ``Optimizing fixation
  prediction using recurrent neural networks for 360$^{\circ }$ video streaming
  in head-mounted virtual reality,'' \emph{IEEE Transactions on Multimedia},
  vol.~22, no.~3, pp. 744--759, 2020.

\bibitem{coors2018spherenet}
B.~Coors, A.~P. Condurache, and A.~Geiger, ``Spherenet: Learning spherical
  representations for detection and classification in omnidirectional images,''
  in \emph{Proceedings of the European Conference on Computer Vision (ECCV)},
  2018, pp. 518--533.

\bibitem{zhu2018prediction}
Y.~Zhu, G.~Zhai, and X.~Min, ``The prediction of head and eye movement for 360
  degree images,'' \emph{Signal Processing: Image Communication}, vol.~69, pp.
  15--25, 2018.

\bibitem{8931644}
Y.~Zhu, G.~Zhai, X.~Min, and J.~Zhou, ``The prediction of saliency map for head
  and eye movements in 360 degree images,'' \emph{IEEE Transactions on
  Multimedia}, vol.~22, no.~9, pp. 2331--2344, 2020.

\bibitem{ZhuYucheng2020Learning}
------, ``Learning a deep agent to predict head movement in 360-degree
  images,'' \emph{ACM Trans. Multimedia Comput. Commun. Appl.}, vol.~16, no.~4,
  Dec. 2020.

\bibitem{petrangeli2017http}
S.~Petrangeli, V.~Swaminathan, M.~Hosseini, and F.~De~Turck, ``An http/2-based
  adaptive streaming framework for 360 \degree virtual reality videos,'' in
  \emph{Proceedings of the 25th ACM International Conference on Multimedia},
  ser. MM '17.\hskip 1em plus 0.5em minus 0.4em\relax New York, NY, USA:
  Association for Computing Machinery, 2017, p. 306–314.

\bibitem{qian2018flare}
F.~Qian, B.~Han, Q.~Xiao, and V.~Gopalakrishnan, ``Flare: Practical
  viewport-adaptive 360-degree video streaming for mobile devices,'' in
  \emph{Proceedings of the 24th Annual International Conference on Mobile
  Computing and Networking}, ser. MobiCom '18.\hskip 1em plus 0.5em minus
  0.4em\relax New York, NY, USA: Association for Computing Machinery, 2018, p.
  99–114.

\bibitem{yadav2020tile}
P.~K. Yadav and W.~T. Ooi, ``Tile rate allocation for 360-degree tiled adaptive
  video streaming,'' in \emph{Proceedings of the 28th ACM International
  Conference on Multimedia}, ser. MM '20.\hskip 1em plus 0.5em minus
  0.4em\relax New York, NY, USA: Association for Computing Machinery, 2020, p.
  3724–3733.

\bibitem{feng2019viewport}
X.~Feng, V.~Swaminathan, and S.~Wei, ``Viewport prediction for live 360-degree
  mobile video streaming using user-content hybrid motion tracking,''
  \emph{Proceedings of the ACM on Interactive, Mobile, Wearable and Ubiquitous
  Technologies}, vol.~3, no.~2, pp. 1--22, 2019.

\bibitem{Wang_2018_CVPR}
W.~Wang, J.~Shen, F.~Guo, M.-M. Cheng, and A.~Borji, ``Revisiting video
  saliency: A large-scale benchmark and a new model,'' in \emph{Proceedings of
  the IEEE Conference on Computer Vision and Pattern Recognition (CVPR)}, June
  2018, pp. 4894--4903.

\bibitem{Jiang_2018_ECCV}
L.~Jiang, M.~Xu, T.~Liu, M.~Qiao, and Z.~Wang, ``Deepvs: A deep learning based
  video saliency prediction approach,'' in \emph{Proceedings of the European
  Conference on Computer Vision (ECCV)}, September 2018, pp. 602--617.

\bibitem{che2019gaze}
Z.~Che, A.~Borji, G.~Zhai, X.~Min, G.~Guo, and P.~Le~Callet, ``How is gaze
  influenced by image transformations? dataset and model,'' \emph{IEEE
  Transactions on Image Processing}, vol.~29, pp. 2287--2300, 2019.

\bibitem{min2016fixation}
X.~Min, G.~Zhai, K.~Gu, and X.~Yang, ``Fixation prediction through multimodal
  analysis,'' \emph{ACM Trans. Multimedia Comput. Commun. Appl.}, vol.~13,
  no.~1, oct 2016.

\bibitem{zhu2021lavs}
D.~Zhu, D.~Zhao, X.~Min, T.~Han, Q.~Zhou, S.~Yu, Y.~Chen, G.~Zhai, and X.~Yang,
  ``Lavs: A lightweight audio-visual saliency prediction model,'' in \emph{2021
  IEEE International Conference on Multimedia and Expo (ICME)}.\hskip 1em plus
  0.5em minus 0.4em\relax IEEE, 2021, pp. 1--6.

\bibitem{eder2019convolutions}
M.~Eder and J.-M. Frahm, ``Convolutions on spherical images,'' in
  \emph{Proceedings of the IEEE/CVF Conference on Computer Vision and Pattern
  Recognition Workshops}, 2019, pp. 1--5.

\bibitem{9303135}
C.~H. Vo, J.-C. Chiang, D.~H. Le, T.~T. Nguyen, and T.~V. Pham, ``Saliency
  prediction for 360-degree video,'' in \emph{2020 5th International Conference
  on Green Technology and Sustainable Development (GTSD)}, 2020, pp. 442--448.

\bibitem{2021Mosaic}
S.~Park, A.~Bhattacharya, Z.~Yang, S.~R. Das, and D.~Samaras, ``Mosaic :
  Advancing user quality of experience in 360-degree video streaming with
  machine learning,'' \emph{IEEE Transactions on Network and Service
  Management}, vol.~PP, no.~99, pp. 1000--1015, 2021.

\bibitem{driscoll1994computing}
J.~R. Driscoll and D.~M. Healy, ``Computing fourier transforms and convolutions
  on the 2-sphere,'' \emph{Advances in applied mathematics}, vol.~15, no.~2,
  pp. 202--250, 1994.

\bibitem{s.2018spherical}
T.~S. Cohen, M.~Geiger, J.~Köhler, and M.~Welling, ``Spherical {CNN}s,'' in
  \emph{International Conference on Learning Representations (ICLR)}, 2018.

\bibitem{lin2011perceptual}
W.~Lin and C.-C.~J. Kuo, ``Perceptual visual quality metrics: A survey,''
  \emph{Journal of visual communication and image representation}, vol.~22,
  no.~4, pp. 297--312, 2011.

\bibitem{Woo_2018_ECCV}
S.~Woo, J.~Park, J.-Y. Lee, and I.~S. Kweon, ``Cbam: Convolutional block
  attention module,'' in \emph{Proceedings of the European Conference on
  Computer Vision (ECCV)}, September 2018.

\bibitem{Zhang_2018_ECCV}
Z.~Zhang, Y.~Xu, J.~Yu, and S.~Gao, ``Saliency detection in 360 \degree
  videos,'' in \emph{Proceedings of the European Conference on Computer Vision
  (ECCV)}, September 2018, pp. 488--503.

\bibitem{8315047}
Z.~Bylinskii, T.~Judd, A.~Oliva, A.~Torralba, and F.~Durand, ``What do
  different evaluation metrics tell us about saliency models?'' \emph{IEEE
  Transactions on Pattern Analysis and Machine Intelligence}, vol.~41, no.~3,
  pp. 740--757, 2019.

\bibitem{liang2020ai}
Q.~Liang, P.~Shenoy, and D.~Irwin, ``Ai on the edge: Rethinking ai-based iot
  applications using specialized edge architectures,'' 2020.

\end{thebibliography}


\end{document}